\documentclass[12pt,english]{article}

\PassOptionsToPackage{unicode}{hyperref}
\PassOptionsToPackage{naturalnames}{hyperref}

\usepackage{algorithm2e}
\RestyleAlgo{ruled}
\SetAlFnt{\small}

\usepackage{mathptmx}
\usepackage[T1]{fontenc}
\usepackage{geometry}
\usepackage{amssymb}
\usepackage{color}
\usepackage{babel}
\usepackage{booktabs}
\usepackage{mathtools}
\usepackage{amsmath}
\usepackage{stmaryrd}
\usepackage{graphicx}
\usepackage{setspace}
\usepackage{esint}
\usepackage{amsmath, amsthm, amssymb}
\newtheorem{prop}{Proposition}
\usepackage{newtxtext,newtxmath}
\usepackage[authormarkup=none,final]{changes}
\definechangesauthor[name={del}, color=red]{del}
\usepackage[unicode=true,pdfusetitle,
bookmarks=true,bookmarksnumbered=false,bookmarksopen=false,breaklinks=false,pdfborder={0 0 1},backref=false,colorlinks=true]
 {hyperref}
\hypersetup{
 linkcolor=blue,urlcolor={blue},filecolor={blue},citecolor={blue}}

\makeatletter

\newcommand{\lyxaddress}[1]{
	\par {\raggedright #1
	\noindent\par}
}

\usepackage{caption}
\usepackage{subcaption}

\usepackage{chngcntr}
\usepackage{mathtools}
\counterwithout{figure}{section}
\counterwithout{equation}{section}
\usepackage{times}
\usepackage{amsmath}
\usepackage{floatrow}	
\usepackage{times}
\usepackage{upgreek}
\usepackage{comment}

\usepackage{floatrow}

\counterwithout{figure}{section}
\counterwithout{equation}{section}
\usepackage{changepage}
\usepackage{babel}

\counterwithout{figure}{section}
\counterwithout{equation}{section}
\usepackage{times}
\usepackage{float}
\usepackage{bm}

\usepackage[numbers,sort&compress]{natbib} 

\date{}
\renewcommand{\fnum@figure}{Fig. \thefigure}
\makeatother
\begin{document}

\title{Multistable Physical Neural Networks}

\author{Ben-Haim Eran, Givli Sefi, Or Yizhar and Gat D.~Amir*}

\maketitle

\lyxaddress{\begin{center}
Faculty of Mechanical Engineering, Technion --Israel Institute of
Technology, Haifa 32000, Israel\\
*Corresponding author: amirgat@technion.ac.il
\par\end{center}}

\begin{abstract}
Artificial neural networks (ANNs), which are inspired by the brain, are a central pillar in the ongoing breakthrough in artificial intelligence. In recent years, researchers have examined mechanical implementations of ANNs, denoted as Physical Neural Networks (PNNs). PNNs offer the opportunity to view common materials and physical phenomena as networks, and to associate computational power with them. In this work, we incorporated mechanical bistability into PNNs, enabling memory and a direct link between computation and physical action. To achieve this, we consider an interconnected network of bistable liquid-filled chambers. We first map all possible equilibrium configurations or steady states, and then examine their stability. Building on these maps, both global and local algorithms for training multistable PNNs are implemented. These algorithms enable us to systematically examine the network's capability to achieve stable output states and thus the network's ability to perform computational tasks. By incorporating PNNs and multistability, we can design structures that mechanically perform tasks typically associated with electronic neural networks, while directly obtaining physical actuation. The insights gained from our study pave the way for the implementation of intelligent structures in smart tech, metamaterials, medical devices, soft robotics, and other fields. 
\end{abstract}

\section{Introduction}
\label{sec:Introduction}

Brains are sophisticated biocomputational systems capable of performing multiple complex tasks simultaneously, yet are robust and remarkably power efficient \cite{lai2006distribution}. Comprising billions of neurons interconnected by trillions of synapses, the human brain achieves these remarkable capabilities through extensive connectivity, a hierarchical functional organization, sophisticated learning rules, and neuronal plasticity. 

Artificial neural networks (ANNs) were developed in order to enable bio-inspired learning capabilities. ANNs, however, are often computed using standard computers and lack many of the advantages of brain computation properties, such as parallelism, low energy consumption, fault tolerance, and inherent robustness \cite{bullmore2012economy}. This led to the development of neuromorphic computing, a field that mimics the brain's computation process using physical mechanisms. The use of spiking neural networks on neuromorphic computing platforms, grounded in neural processing principles, provides researchers with valuable insights for constructing adaptive and efficient artificial intelligence (AI) systems \cite{burr2017neuromorphic,markovic2020physics}. This approach, influenced by advancements in materials engineering, device physics, chip integration, and neuroscience, has generated significant interest from both neuroscientists and computer scientists.

Comprehensive reviews, published in recent years, explored various facets of neuromorphic computing, covering device physics \cite{duan2020spiking,gw2020hwang,kuzum2013synaptic}, circuit design \cite{indiveri2011neuromorphic,bartolozzi2007synaptic}, and network integration \cite{roy2019towards,furber2016large}. Training of neuromorphic systems, associated with physical learning, requires the modification of physical elements to provide desired computational outcomes. Previous studies involved training both simulated and laboratory mechanical networks, along with simulated flow networks, to perform specific tasks by adjusting their internal degrees of freedom \cite{kendall2020training,stern2023learning,pashine2019directed,pashine2021local,scellier2021deep,stern2021supervised,goodrich2015principle,rocks2017designing,stern2018shaping,rocks2019limits,ruiz2019tuning,hexner2020periodic,stern2020continual}. This training is achieved through the minimization of a global cost function \cite{stern2021supervised,goodrich2015principle,rocks2017designing,stern2018shaping,rocks2019limits,lee2022mechanical} or the application of local rules facilitated by an external processor \cite{kendall2020training,pashine2019directed,pashine2021local,scellier2021deep,ruiz2019tuning,hexner2020periodic,stern2020continual}.

The scope of learning capacity extends beyond the limits of the brain; even muscles have been observed learning and making decisions in organisms such as ciliates \cite{bull2021ciliary}, fruit flies \cite{ristroph2013active}, and bee hives \cite{peleg2018collective}. Additionally, understanding learning principles can be harnessed for accomplishing intricate tasks such as data classification \cite{stern2020continual,stern2021supervised,dillavou2022demonstration,anisetti2023learning} or regression \cite{dillavou2022demonstration}. Mechanical \cite{stern2020continual,stern2021supervised}, flow \cite{stern2020continual,anisetti2023learning}, and resistor networks \cite{dillavou2022demonstration} can perform these tasks by transforming data into physical stimuli and responses, aligning with the common applications of machine learning. Moreover, these systems, devoid of a processor or memory storage, exhibit high robustness to damage, suggesting a novel direction for computer design. The interplay between learning and memory in physical systems underscores the importance of understanding properties contributing to memory retention, paving the way for leveraging material memory capabilities in training applications \cite{crowder2014principles,anderson2000learning,stern2023learning}.

The development of metamaterials with self-learning capabilities opens possibilities for creating intelligent metamaterials \cite{ma2019smart,stern2023learning}.  
Moreover, metamaterials such as mechanical neural networks are being explored as potential solutions for precise control in soft robotics involving bistable elements. 
Bistability streamlines control complexities, offering energy-efficient and adaptable solutions for reliable soft robotic systems \cite{gorissen2019hardware,gorissen2020inflatable,ben2020single,cao2021bistable}. The dual-stable nature of bistable soft actuators conserves energy in static positions and enhances adaptability for variable stiffness and shape changes \cite{melancon2022inflatable}. Bistable systems provide memory, programmability, and compact design, making them suitable for diverse applications, including wearable robotics and medical devices \cite{chi2022bistable,receveur2005laterally}.

In this research, \textit{we aim to incorporate mechanical bistability into a physical neural network }(PNNs), enabling memory and a direct link between computation and physical action.

The physical neural network examined in this work is constructed from a flow network characterized by bistable nodes, which incorporate internal, external, and output nodes. 
Each bistable element within the network exhibits a non-linear pressure-volume relationship, characterized by two stable equilibrium states of positive stiffness, identified as binary states `0' and `1'. The objectives of this study are: firstly, to delineate the potential equilibrium configurations or the steady states achievable by the network; secondly, to conduct an in-depth examination of their stability; and finally, to train the network to perform desired tasks. To this end, we propose two tailored methodologies for analyzing bistable flow networks, each designed to meet the unique requirements of the system under study. The first method investigates the behavior of the network in the absence of external pressure forces; the system is capable of manifesting $2^{N}$ distinct binary states in an equilibrium state. 
The core challenge involves determining an optional network topology (and resistance configuration) that drives the system toward a desired equilibrium state within the array of possible states. In order to address this challenge, we employ a global supervised learning strategy.
The analysis then pivots to a second methodology, where the input nodes are subjected to predefined boundary conditions and pressures. The aim here is to investigate the network's parameters such that the output nodes' pressures and binary states precisely match the desired targets. This is achieved through the adoption of a local physical supervised learning approach.
Adaptation in the learning process leverages the mechanical properties of the network's elements, such as the stiffness or equilibrium positions of springs in a mechanical context, or the conductances in a flow network, thereby providing the network with the capacity to evolve towards specified operational goals.

\section{Formulation of flow networks with bistable chambers}

This work aims to combine multistability into physical neural networks. There are various possible physical systems, and in this study, we will focus on a physical network consisting of $N$ interconnected hyperelastic chambers (being the nodes) linked by rigid tubes (see Fig.\,\ref{fig: Illustration}).
This section models the bistable flow network, maps equilibrium and steady state configurations, and analyzes the stability of the network. 

\subsection{Modeling}
The volumetric flow rates, $Q_{ij}$, between the $i^{th}$ and $j^{th}$ chambers, governed by $Q_{ij}=C_{ij}(p_{j}-p_{i})$, exhibit a linear dependence on their conductances, $C_{ij}$, where $C_{ij}=1/R_{ij}$ and $R_{ij}=8\mu\ell_{ij}/\pi a_{ij}^{4}$ represents the viscous resistance of the tube, and $p_{i}(t)$ is the hydrostatic pressure inside the $i^{th}$ chamber. This dependency is influenced by the tube's length, $\ell_{ij}$, and is inversely proportional to the fourth power of the tube radius, $a_{ij}$, following the Hagen–Poiseuille equation governing laminar fluid flow \cite{pfitzner1976poiseuille}. Note that the fluid viscosity, $\mu$, is assumed to be constant. Each node in the network can undergo volume changes, denoted by $v_{i}(t)$. 
To explore the flow dynamics within the network, we formulate the system equations in matrix form. The pressures vector is represented as $\mathbf{p}(t)=[p_{1},\cdots,p_{N}]^{T}$, and the nodes are modeled as elastic bodies (i.e., chambers) with volumes $\mathbf{v}(t)=[v_{1},\cdots,v_{N}]^{T}$. We assume hydrostatic pressure within nodes \cite{ben2022viscous,beatty1987topics}. Furthermore, it is assumed that the viscous resistance of the nodes is negligible compared to that of the tubes. The external volumetric flow is expressed as $\mathbf{q}(t)=[q_{1},\cdots,q_{N}]^{T}$, where $q_{i}$ represents a prescribed external volumetric flow into the $i^{th}$ node.

Mass conservation at each node, while considering the entire network's conductances, yields
\begin{equation}
\label{eq: Equation of motion}
    \frac{\mathrm{d}\mathbf{v}}{\mathrm{d}t}=-\mathbf{W}\mathbf{p}+\mathbf{q},
\end{equation}
which is equivalent to Kirchhoff’s law for the current system. The expression \(W_{ij} = -C_{ij} + \delta_{ij}\sum_{k}C_{ik}\) represents the (weighted) graph Laplacian matrix, where $\delta_{ij}$ denoted Kronecher delta. It takes into account not only the network's connection topology but also the tubes' viscous resistance. The matrix $\mathbf{W}$ is singular, with all non-zero eigenvalues having a positive real part. We assume that the physical network can be represented as a connected graph, and that $\mathbf{W}$ is irreducible and its rank is $N-1$, with the smallest eigenvalue being zero and all others positive. Furthermore, $\text{Ker}(\mathbf{W})$ is spanned by $\mathbf{1}_{N}$ where $\mathbf{1}_{N}$ denotes the $N$-vector of all unit elements, and $\text{Im}(\mathbf{W})=\mathbf{1}_{N}^{\perp}$, where $\mathbf{1}_{N}^{\perp}$ is the subspace defined by $\mathbf{1}_{N}^{\perp}=\{\mathbf{x}\in\mathbb{R}^{N}\,\big|\,\mathbf{1}_{N}^{T}\mathbf{x}=0\}$. 
Comprehensive mathematical details, including theorems and supplementary proofs, can be explored in classical literature on graph theory, such as reference \cite{bullo2020lectures}.
The $N$ nodes of the network include three subsets:  $b$ external inlet nodes (denoted by $\mathcal{B}_{0}$) that serve as boundary conditions, $t$ output nodes (denoted by $\mathcal{T}_{0}$) that serves as targets node. The remaining $d$ internal nodes are referred to as "hidden" nodes (denoted by $\mathcal{D}_{0}$). We note that $d+b+t=N$ holds.
In addition, we divide the external nodes into two distinct types of boundary conditions: pressure-constrained external nodes and volumetric-constrained external nodes. 

\begin{figure}[t!]
    \centering
    \includegraphics[width=1\textwidth]{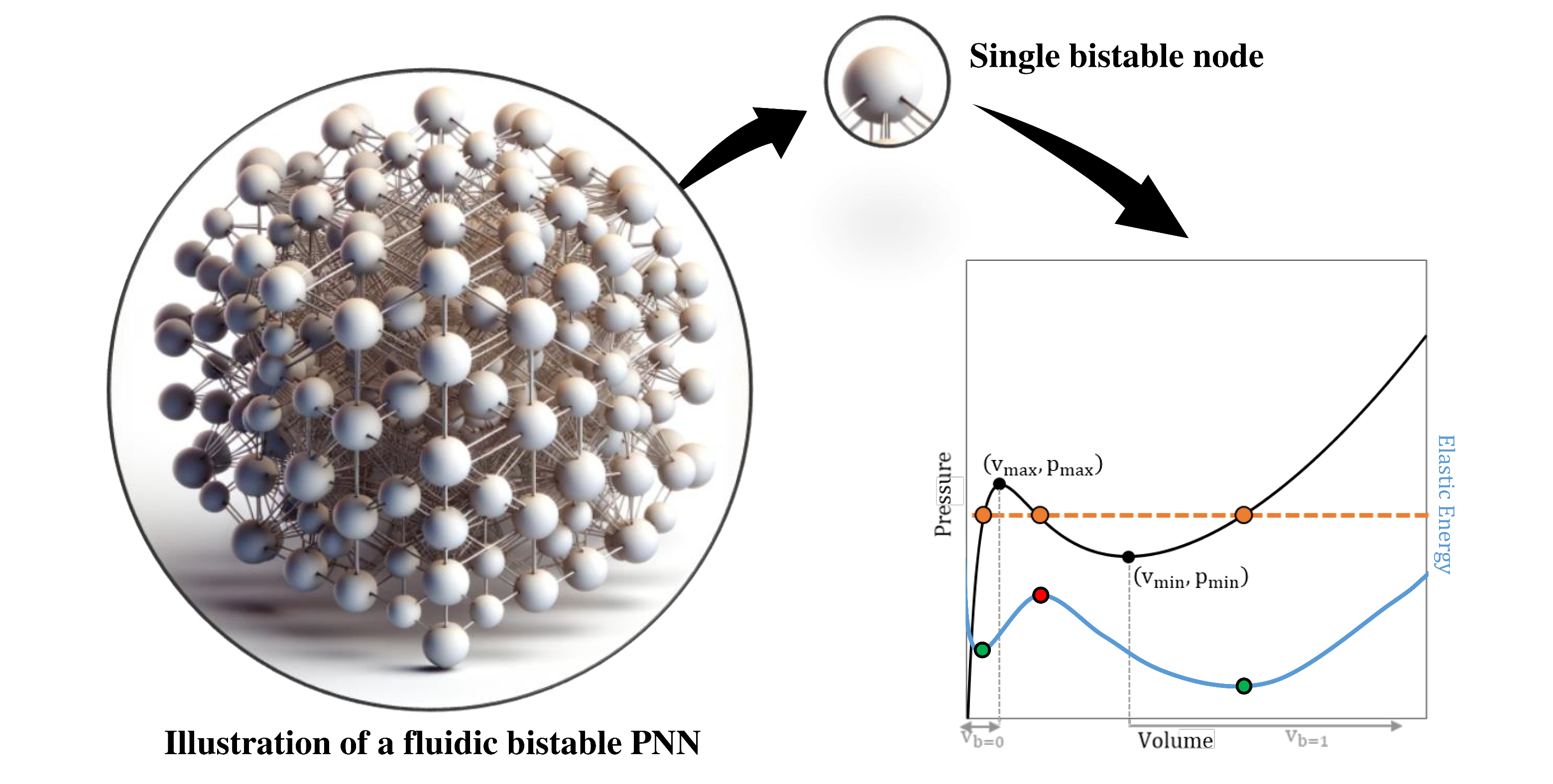}
    \caption{Illustration of a hierarchical metamaterial structure composed of bistable nodes interconnected by rigid tubes, showcasing the concept of design of the physical neural network in advanced metamaterial framework (image generated by OpenAI's DALL-E). Each node within this network is characterized by a distinct non-monotonic pressure-volume relationship, leading to two practical and stable equilibrium phases, symbolized as $v_{b=0}$ and $v_{b=1}$, or more simply, `0' and `1'. Illustrated by the blue curve is the node's elastic potential, which clearly delineates two stable states separated by an unstable state, associated with the spinodal branch.}
    \label{fig: Illustration}
\end{figure}

We define a functional relationship between node pressure and volume, expressed as \(\mathbf{p} = \mathbf{\nabla}^{T}_{\mathbf{v}}\mathbf{\psi}\), where $\mathbf{\psi}$ represents the total elastic energy, given by $\psi(\mathbf{v})=\psi_{1}(v_{1})+\cdots+\psi_{N}(v_{N})$. This functional relation is applicable in situations where a (quasi-)static equilibrium is maintained, with the internal pressure of the body balanced by structural forces that counteract the pressure. Several examples of the use of such relations can be found in \citet{beatty1987topics} and \citet{,dal2023silico}.

In the linear framework, \(\mathbf{p}= \sigma \mathbf{v}\) for a classical setting, where \(\sigma\) represents a known constant factor. However, this relationship deviates from linearity and relies on distinct governing laws in scenarios involving multistable bodies such as bistable elements.
A \textit{bistable} element is characterized by two distinct stable equilibrium configurations under the same prescribed load, separated by an unstable (spinodal) equilibrium state, as illustrated in Fig.\,\ref{fig: Illustration} in the pressure-volume characterization.
Each bistable element is defined by a non-monotonic pressure-volume relation, $p_{i}=f(v_{i})$, featuring two phases of positive stiffness, separated by an intermediate branch characterized by negative stiffness.
For convenience, it is assumed that all elastic bodies have the same pressure-volume relation.
Consequently, two distinctly separated equilibrium phases are denoted as `0' and `1' associated with volumes $v\in v_{b=0}$ for the `0' binary state, or $v\in v_{b=1}$ for the `1' binary state. The local minimum and maximum points of the pressure-volume relation are denoted as $(v_{min},p_{min})$ and $(v_{max},p_{max})$, respectively.
Such behavior is observable in various structures such as curved beams \citep{salinas2015can,vangbo1998analytical,zhao2008post,mises1923stabilitatsprobleme,arena2017adaptive,timoshenko2009theory}, thin-walled hyper-elastic balloons \citep{alexander1971tensile,muller2004rubber,overvelde2015amplifying,ben2020single}, and pre-stressed elastic sheets \citep{brodland1987deflection,faber2020dome,liu2023snap,shui2022aligned}.

\label{sec: Flow networks with bistable nodes}

\subsection{Network equilibrium and steady states}
We define the initial volume of fluid within the network by $v_{0}$, which includes the chambers and tubes. To assess the total volume of the flow as a function of time, we consider the set of all nodes as a control volume, i.e.,
\begin{equation}
\label{eq: v}
    \mathbb{V}(t)=v_{0}+\int_{0}^{t}{\mathbf{1}_{N}\cdot\mathbf{q}}(\tau)\mathrm{d}\tau.
\end{equation}
Steady states of the system, where the total volume is conserved as $\mathbb{V}_{ss}=\lim_{t\rightarrow\infty}{\mathbb{V}(t)}$, necessitate that all nodes' pressures remain constant, namely, $\mathbf{W} \mathbf{p}_{ss}=\mathbf{q}_{ss}$, where $\mathbf{q}_{ss}$ denoted the flux injections vector where $t\gg 1$. Consequently, the latter equation has a solution if and only if the flux injections are balanced at steady state, which means that $\mathbf{q}_{ss}\in\mathbf{1}_{N}^{\perp}$.

In cases in which the external nodes are solely influenced by the prescribed flow rate, the steady state solution is given by
\begin{equation}
\label{eq: p_ss}
    \mathbf{p}_{ss}=\alpha\cdot\mathbf{1}_{N}+\mathbf{W}^{\dag}\mathbf{q}_{ss},
\end{equation}
with $\alpha\in\mathbb{R}$. The first right-hand term corresponds to the homogeneous solution, signifying a scenario with no injections into the network and equal pressures along the network (i.e., equilibrium state). The second right-hand term represents the particular solution, where $\mathbf{W}^{\dag}$ is the \textit{Moore-Penrose} inverse of the Laplacian matrix \cite{meyer2023matrix}.

In scenarios where a subset of $1 \leq b_{1}\leq b$ external nodes undergo pressure constraints, $\mathbf{p_{BC}}\in\mathbb{R}^{b_{1}\times 1}$ (including the constrained unknown volumetric flux, $\mathbf{q_{C}}\in\mathbb{R}^{b_{1}\times 1}$), while the remaining $b_{2}=b-b_{1}$ nodes are governed by known volumetric fluxes, $\mathbf{q_{BC}}\in\mathbb{R}^{b_{2}\times 1}$; the system converges on a unique solution set that satisfies the boundary conditions. In order to determine the steady state solution, the first step involves algebraic reconfigurations of the steady state equations. These manipulations aim to convert the equations into a structured form:
\begin{equation}
\label{eq: p_ss2}
\left[ 
\begin{array}{c|c|c} 
  \mathbf{\hat{W}_{11}}\in\mathbb{R}^{b_{1}\times b_{1}}\quad & \mathbf{\hat{W}_{12}}\in\mathbb{R}^{b_{1}\times b_{2}}\quad & \mathbf{\hat{W}_{13}}\in\mathbb{R}^{b_{1}\times (N-b)}\quad \\ 
  \hline 
  \mathbf{\hat{W}_{21}}\in\mathbb{R}^{b_{2}\times b_{1}}\quad &  \mathbf{\hat{W}_{22}}\in\mathbb{R}^{b_{2}\times b_{2}}\quad & \mathbf{\hat{W}_{23}}\in\mathbb{R}^{b_{2}\times (N-b)}\quad \\
  \hline
  \mathbf{\hat{W}_{31}}\in\mathbb{R}^{(N-b)\times b_{1}} & \mathbf{\hat{W}_{32}}\in\mathbb{R}^{(N-b)\times b_{2}} & \mathbf{\hat{W}_{33}}\in\mathbb{R}^{(N-b)\times (N-b)}
\end{array} 
\right]
\begin{bmatrix}
  \mathbf{p_{BC}}\\
  \cmidrule(lr){1-1}
  \mathbf{p_{ss}^{ext}}\\
  \cmidrule(lr){1-1}
  \mathbf{\bar{p}_{ss}}
\end{bmatrix}
=
\begin{bmatrix}
  \mathbf{q_{C}}\\
  \cmidrule(lr){1-1}
  \mathbf{q_{BC}}\\
  \cmidrule(lr){1-1}
  \mathbf{0}
\end{bmatrix}
\end{equation}
where $\mathbf{p_{ss}^{ext}}$ is the steady state pressure vector of the $b_{2}$ external nodes which is constrained by the known flux, $\mathbf{q_{BC}}$. Following Eq.\,(\ref{eq: p_ss2}), the steady state solution for the unforced nodes is obtained 
\begin{equation}
\label{eq: p_ss_solution}
    \mathbf{\bar{p}}_{ss}=\big(\mathbf{\hat{W}_{32}}\mathbf{\hat{W}_{22}^{-1}}\mathbf{\hat{W}_{23}}-\mathbf{\hat{W}_{33}}\big)^{-1}\big[\big(\mathbf{\hat{W}_{31}}-\mathbf{\hat{W}_{32}}\mathbf{\hat{W}_{22}^{-1}}\mathbf{\hat{W}_{21}}\big)\mathbf{p_{BC}}+\mathbf{\hat{W}_{32}}\mathbf{\hat{W}_{22}^{-1}}\mathbf{q_{BC}}\big].
\end{equation}

We thus observe that, in the absence of external forcing (i.e., a closed system without flux or prescribed pressure), the system displays uniform pressure across all nodes upon reaching an equilibrium state. Conversely, in scenarios where the system is subjected to external forces (be it through flux, pressure, or a combination of these forces) the resultant pressures at the nodes, post convergence (i.e., in steady state), are constituted by a linear combination of the system's constraints. This linear combination is significantly influenced by the elements of the Laplacian matrix, indicative of the system's dependence on both the network topology and the viscous resistances of the tubes. While it might be assumed that architecting a network with a specific configuration could engineer a transformation conducive to a predefined equilibrium configuration or a specific steady state, complexity arises when the quantity of tubes exceeds the number of nodes within a given set of constraints. This gap renders the transformation indeterminate, lacking a single-valued definition. Consequently, it remains unclear whether a consistent and desirable transformation can always be established between a set of constraints and steady states. This uncertainty highlights the complicated relationship between network architecture and its operational dynamics.

Once the equilibrium or the steady states of the system have been identified, the discussion turns to exploring the implications of setting a specific set of system pressures on the corresponding nodal volumes. 
The relationship between pressure and volume at the nodes is crucial for understanding the system's dynamics, as it directly impacts flow rates, system capacity, and operational efficiency. For linear relationship, the combination of the known total volume, $\mathbb{V}_{ss}$, with the solution provided in Eq.\,(\ref{eq: p_ss}) or Eq.\,(\ref{eq: p_ss_solution}) yields a single solution, delineated by the steady state configuration of the network. However, in the context of nonlinear nodes (i.e., non linear relationship between pressure and volume), such as multistable bodies, the aforementioned conditions may yield multiple solutions for node volumes. Since bistable relationships are not inherently reversible, the system's state is not uniformly defined in certain scenarios, so to extract volumes from pressure information, it is necessary to model the states of bistable chambers.

\subsection{Network stability}
\label{subsection: Network equilibrium stability}
In this section, we analyze the stability of equilibrium states and steady states of the network. The elastic network is governed by a scalar elastic energy function, $\psi_{i}(v_{i})$, and the total elastic energy of the nodes is $\psi(\mathbf{v})=\psi_{1}(v_{1})+\cdots+\psi_{N}(v_{N})$. 
The total volume of flow within the network is known, according to Eq.\,(\ref{eq: v}). It is not necessary to solve all $N$ equations of motion. The resolution of $N-1$ equations is sufficient, achieved by removing one node arbitrarily. This degree of freedom is determined by knowing the total volume of the system.

Without loss of generality, let's eliminate the equation of motion for the $N^{th}$ node, assuming this particular node is not included in the external inlet set (i.e. the BCs). Returning to the general dynamic equation (\ref{eq: Equation of motion}), we omit the $N^{th}$ degree of freedom, excluding the $N^{th}$ row and column from the Laplacian matrix, along with the $N^{th}$ element in the vectors $\mathbf{q}$, $\mathbf{p}$ and $\mathbf{v}$. The reduced system can be expressed as $\mathrm{d}\tilde{\mathbf{v}}/\mathrm{dt}=-\mathbf{\tilde{W}\tilde{p}}+\mathbf{\tilde{q}}$ subjected to effective constraint (\ref{eq: v}).   
The constrained elastic energy function can be eliminated as $\tilde{\psi}(\mathbf{\tilde{v}})=\psi\big(\mathbf{\tilde{v}},v_{N}(\mathbf{\tilde{v}})\big)$ where $\tilde{\mathbf{v}}=[v_{1},\cdots,v_{N-1}]^{T}$ is the reduced state vector, and $v_{N}(\mathbf{\tilde{v}})=\mathbb{V}-(v_{1}+\cdots+v_{N-1})$.

The subsequent sections focus on a formalized criterion for stability, supported by formal proofs, which are detailed in the appendix \ref{Proofs}. This approach improves the understanding of the system's dynamics and provides a robust framework for assessing stability.

\begin{prop}
\label{prob:Reduced EOM}
For the constrained system, the (reduced) system can be described by 
\begin{equation}
\label{eq: constrained equation of motion}
    \frac{\mathrm{d}\tilde{\mathbf{v}}}{\mathrm{dt}}=-\mathbf{\tilde{W}\nabla_{\mathbf{\tilde{v}}}^{T}\tilde{\psi}}+\mathbf{\tilde{q}},
\end{equation}
where $\mathbf{\nabla_{\mathbf{\tilde{v}}}^{T}\psi(\mathbf{\tilde{v}})}$ is the gradient of the constrained elastic energy function with respect to the reduced state vector.
\end{prop}

\begin{prop}
\label{prob:Reduced W}
The reduced matrix $\mathbf{\tilde{W}}$ possesses square, real, symmetric, and positive definite properties. 
\end{prop}

The Hessian matrix of the constrained elastic energy function, denoted as $\mathbb{H}_{\mathbf{\tilde{v}}}(\tilde{\psi})$, plays a pivotal role in understanding the dynamic stability of the network. However, it's critical to note from the dynamic equations outlined in Eq.\,(\ref{eq: constrained equation of motion}) that the elastic energy $\tilde{\psi}$ cannot be directly equated to the total effective potential energy of the system. This is attributed to the fact that the time derivative of the state vector is influenced not merely by the gradient of $\tilde{\psi}$ but by a specific combination of gradient elements through multiplication in matrix $\mathbf{\tilde{W}}$. Consequently, the stability of the system cannot be directly determined from the elastic energy $\tilde{\psi}$.
Nevertheless, we present a methodology where a special transformation from the original state vector of the system (represented by coordinate set $\mathbf{\tilde{v}}$) to an alternative state vector (represented by the coordinate set $\mathbf{\tilde{s}}$) is feasible, enabling the definition of a scalar function representing the effective potential energy of the system. 

\begin{prop}
\label{prob:Reduced T}
There exists a regular matrix $\mathbf{T}\in\mathbb{R}^{(N-1)\times(N-1)}$ such that $\mathbf{T}\mathbf{\tilde{W}}\mathbf{T}^{T}=\mathbf{I}$, where $\mathbf{I}$ is the identity marix.
\end{prop}

Since $\mathbf{T}$ is regular, it can be used as a linear transformation from vector coordinate $\mathbf{\tilde{v}}$ to another vector coordinate $\mathbf{\tilde{s}}$, denoting by $\mathbf{\tilde{s}}=\mathbf{T}\mathbf{\tilde{v}}$.

\begin{prop}
\label{prob:Reduced EOM s}
The governing equations of the reduced system can be formulated by the transformed coordinates as $\mathrm{d}\mathbf{\tilde{s}}/\mathrm{dt}=-\nabla_{\mathbf{\tilde{s}}}^{T}\tilde{\psi}(\mathbf{\tilde{s}})+\mathbf{\tilde{g}}$, where $\tilde{\psi}(\mathbf{\tilde{s}})=\tilde{\psi}(s_{1},\cdots,s_{N-1})$ and $\mathbf{\tilde{g}}=\mathbf{T}\mathbf{\tilde{q}}$.
\end{prop}

\begin{prop}
\label{prob_effectiv}
There exist an effective potential energy function, denoted by $\tilde{\phi}(\mathbf{\tilde{s}})=\tilde{\psi}(\mathbf{\tilde{s}})-\mathbf{\tilde{g}}^{T}\mathbf{\tilde{s}}$, such that $\mathrm{d}\mathbf{\tilde{s}}/\mathrm{dt}=-\nabla_{\mathbf{\tilde{s}}}^{T}\tilde{\phi}(\mathbf{\tilde{s}})$.
\end{prop}

\begin{prop}
\label{prob_ss}
Invariant equilibrium points exist between the system formulated by $\mathbf{\tilde{v}}$ and the system formulated by $\mathbf{\tilde{s}}$. In other words, $\mathrm{d}\mathbf{\tilde{v}}/\mathrm{dt}=\mathbf{0}$ if and only if $\mathrm{d}\mathbf{\tilde{s}}/\mathrm{dt}=\mathbf{0}$, and if and only if  $\nabla_{\mathbf{\tilde{s}}}^{T}\tilde{\phi}(\mathbf{\tilde{s}})=\mathbf{0}$.
\end{prop}

The outcome derived from propositions \ref{prob_effectiv} and \ref{prob_ss} suggests that characterizing the stability of equilibrium states necessitates ensuring the Hessian matrix of the effective potential energy, denoted as $\mathbb{H}_{\mathbf{\tilde{s}}}(\tilde{\phi})$,  is positive definite, which means all its eigenvalues should be positive.

\begin{prop}
\label{prob:similarity}
The Hessian matrices of the reduced potential energy $\mathbb{H}_{\mathbf{\tilde{s}}}(\tilde{\phi})$ and the (constrained) elastic energy $\mathbb{H}_{\mathbf{\tilde{v}}}(\tilde{\psi})$, are similar.
\end{prop}
Remarkably, the Hessian matrix concerning the effective energy—derived via second derivatives with respect to the coordinates $\mathbf{\tilde{s}}$—bears similarity to the Hessian matrix of the elastic energy function, $\mathbb{H}_{\mathbf{\tilde{v}}}(\tilde{\psi})$. This similarity extends to their eigenvalues being identical. Thus, by examining the eigenvalues of $\mathbb{H}_{\mathbf{\tilde{v}}}(\tilde{\psi})$ one can infer the stability of the system's equilibrium points.
Consequently, assessing the stability of the equilibrium states merely necessitates the Hessian matrix of the (constrained) elastic energy being positive defined, i.e., $\mathbb{H}_{\mathbf{\tilde{v}}}(\tilde{\psi})\succ 0$.
To ensure positive definiteness, we require that the principle minors of the Hessian should be positive, namely,

\begin{equation}
\label{eq: minors}
    \mathbb{M}_{y}=\bigg[\prod_{i=1}^{y} {\cfrac{\mathrm{d}f_{i}}{\mathrm{d}v_{i}}}\bigg]\bigg[1+\cfrac{\mathrm{d}f_{N}}{\mathrm{d}v_{N}}\sum_{i=1}^{y}{\bigg(\cfrac{\mathrm{d}f_{i}}{\mathrm{d}v_{i}}\bigg)^{-1}}\bigg]>0,\quad\text{for all}\quad y\in[1,\cdots,N-1].
\end{equation}
As discussed in \cite{puglisi2000mechanics}, the equilibrium state of the network is based upon the distribution of nodes within the spinodal branch. A network devoid of nodes in the spinodal branch is in a state of stable equilibrium. Conversely, the existence of two or more nodes within the spinodal branch signifies an unstable equilibrium for the network. In the case where only a single node lies in the spinodal branch, the network achieves stability solely under the condition that 
\begin{equation}
\label{eq: stable1spinodal}
    \sum_{i=1}^{N}{\bigg(\cfrac{\mathrm{d}f_{i}}{\mathrm{d}v_{i}}\bigg)^{-1}}<0.
\end{equation}

The study of the network's steady states and the stability of its states reveals that in the absence of external forces, the network will attain a pressure equilibrium across any arbitrary topology. When the system is subjected to external forces, the steady state pressures achieved are influenced by both the network's topology and the resistances of the tubes. While the existence of steady state may rely on the topology and resistances, the steady state stability of the network remains unaffected by these parameters. This finding catalyzes further investigation into bistable networks.

\subsection{Resistance tuning for binary state convergence in a four-node network}
\label{subsub: Resistance investigation}
\begin{figure}
    \centering
    \includegraphics[width=0.9\textwidth]{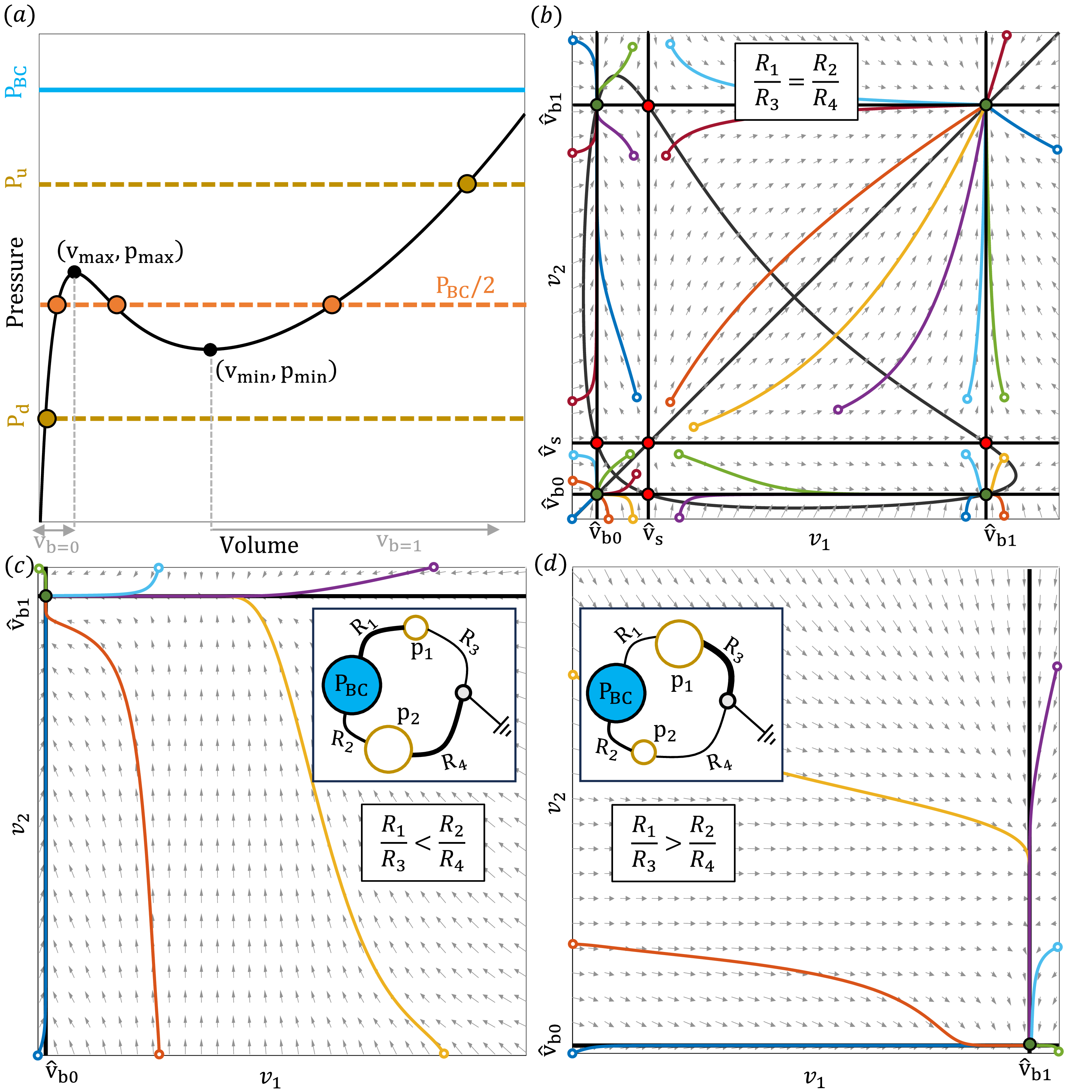}
    \caption{A theoretical investigation of a grounded flow network with four bistable nodes dictated by external pressure $p_{BC}$. The viscous resistances are denoted by $R_{1,2,3,4}$. 
    (a) Typical pressure-volume curve of a bistable element. The domain of volumes in binary states '0' and '1' is $v\in v_{b=0}$ and $v\in v_{b=1}$, respectively. A pressure higher than $p_{max}$ or lower than $p_{min}$ is denoted by $p_{u}$ and $p_{d}$, respectively, and the volumes are defined in a one-value manner. For a pressure within the bistable domain, there are three different options for volume, marked by orange dots.
    (b) $\{v_{1};v_{2}\}$ space to describe the dynamic solutions of the system for the case where $R_{1}/R_{3}=R_{2}/R_{4}$. The black lines are solutions of the steady state equations. The intersection between these lines describes the fixed points of the system (green-stable, red-unstable). The gray arrows describe the dynamic solution field, namely, $\{v_{1}(t),v_{2}(t)\}$. Several solution trajectories are presented for different initial conditions (marked with empty circles). The colors of the curves have no meaning. 
    (c) $\{v_{1};v_{2}\}$ space to describe the network dynamics for the case where $R_{1}/R_{3}<R_{2}/R_{4}$. In the described map, there is one fixed point described by the stable binary state (0,1) to which the network reaches from any initial conditions.
    (d) $\{v_{1};v_{2}\}$ space to describe the network dynamics for the case where $R_{1}/R_{3}>R_{2}/R_{4}$. In the described map, there is one fixed point described by the stable binary state (1,0) to which the network reaches from any initial conditions.}
    \label{fig:Resistences}
\end{figure}

In this section, we analyze a simple network consisting of four elements as a preliminary case study (see the inset in Fig.\,\ref{fig:Resistences}), motivating the overarching goal of guiding flow networks toward a specific equilibrium state with known inputs. 
A pressure $p_{BC}$ is prescribed to the inlet node; another non-adjacent node is grounded (i.e., zero pressure). The viscous resistances of the tubes are indicated by $R_{1,2,3,4}$, with $R_{1}$ and $R_{2}$ representing the resistances between the inlet node and the nodes number 1 and 2, respectively; and $R_{3}$ and $R_{4}$ representing the resistance between the nodes number 1 and 2 and the grounded node.
At steady state, according to Eq.\,(\ref{eq: p_ss_solution}), the pressures of the two outlet node satisfy
\begin{equation}
\label{eq: p1p2ss}
    p_{1}= \frac{1}{1+R_{1}/R_{3}}p_{BC}, \qquad p_{2}= \frac{1}{1+R_{2}/R_{4}}p_{BC}.
\end{equation}
It is observed that the nature of the steady state, as described in Eq.\,(\ref{eq: p1p2ss}), is dictated by the ratios of viscous resistances, namely, $R_{1}/R_{3}$ and $R_{2}/R_{4}$. To illustrate the system's behavior, our analysis is delineated into two distinct scenarios—the first involves identical ratios. Subsequently, we explore the effects of different ratios.

In the scenario where the ratios satisfying $R_{1}/R_{3} = R_{2}/R_{4}$, it is clearly apparent that steady state is achieved when the pressures at the nodes equalize among themselves and $p_{1}=p_{2}=p_{BC}/2$.
Considering this steady state equation, when $p_{BC} > 2p_{max}$ or $p_{BC} < 2p_{min}$, a unique set of nodes' volumes is observed, as depicted in Fig.\,\ref{fig:Resistences}(a) by dark-yellow points. This prescribed pressure leads to a stable binary configuration of (1,1) for the former case and (0,0) for the latter. Here, the binary state of node number 1 is denoted first, followed by the binary state of node number 2.
However, the steady state equation may describe multiple solutions for the nodes' volumes, if $2p_{max} < p_{BC} < 2p_{min}$. Under such conditions, each node may take one of three potential options, represented as orange points in Fig.\,\ref{fig:Resistences}(a). Consequently, we identify nine possible binary configurations, namely, (0,0), (0,s), (0,1), (1,0), (1,s), (1,1), (s,0), (s,s), and (s,1), where `s' represents the spinodal state. 
To examine the network's nature and the stability of resultant steady states, especially under conditions presenting multiple steady state possibilities, we direct the reader's attention to Fig.\,\ref{fig:Resistences}(b). This figure illustrates, within the state space $\{v_1;v_2\}$, the steady state curves alongside the system's dynamic solutions under varying initial conditions. The solutions to the initial steady state equation $p_1 = p_2$ are shown in the black curve, which includes a trivial line $v_1 = v_2$ and two additional curves forming a triangle-like feature for non-trivial solutions, where $v_1 \neq v_2$.
Furthermore, the steady state must also satisfy $p_{BC} = 2p_1$ and $p_{BC} = 2p_2$. These conditions are represented by three distinct lines within the $\{v_1;v_2\}$ space, each illustrating the three potential volumes for pressures that lie between the local minimum and maximum pressure points in pressure-volume characteristic —$\hat{v}_{b0}\in v_{b=0}$ for volumes within the binary domain '0', $\hat{v}_{b1}\in v_{b=1}$ for the binary domain '1', and $\hat{v}_{s}$ for the spinodal domain. Consequently, by intersecting all steady state curves within this space, nine fixed points emerge. Points of stable configuration are highlighted in green, whereas unstable configuration points are denoted in red. The stability of these points is determined based on the analysis section, \ref{subsection: Network equilibrium stability}.

In the scenario where $R_{1}/R_{3} \neq R_{2}/R_{4}$, the equilibrium pressures between the two nodes diverge, breaking the symmetry between the network's branches. Through a strategic choice of resistance ratios, the network can achieve two stable but opposite binary states under the same applied pressure.
For illustrative purposes, we assume at steady state one node stabilizes at a pressure above $p_{max}$ (denoted as $p_{u}$, illustrated in Fig.\,\ref{fig:Resistences}(a)), while the pressure at the other node is below $p_{min}$, represented as $p_{d}$. This configuration yields a single steady state, contingent upon the resistance ratios. If $R_{1}/R_{3} > R_{2}/R_{4}$, it follows that $p_{1} = p_{u}$ and $p_{2} = p_{d}$, guiding the network to a stable steady state in the (1,0) binary configuration from any initial condition. Conversely, if $R_{1}/R_{3} < R_{2}/R_{4}$, then $p_{1} = p_{d}$ and $p_{2} = p_{u}$, resulting in a stable steady state in the (0,1) binary configuration from any initial state. These outcomes are displayed in Fig.\,\ref{fig:Resistences}(c)-(d).
It is relevant to note that the removal of the assumption regarding the nodes' pressures relative to $p_{max}$ and $p_{min}$ introduces additional stable binary states, such as (0,0) or (1,1). In this case, the system's convergence towards a specific state becomes dependent on the initial conditions, further enriching the dynamical behavior.

From the above, it is evident that both the network topology as well as the tube resistances have a significant influence on the network's steady state, specifically regarding the binary state configuration, in terms of convergence of flow. While the topology does not alter the stability of the steady states, it impacts the existence of these points and the network's convergence basins towards equilibrium.
Given a predetermined input, planning of topologies can steer networks towards specified steady states. However, the scenario analyzed here represents a basic and degenerate scenario. The structure of networks with more interconnections and complexity, on the other hand, requires more advanced methods. This study will elaborate on these methods in subsequent sections.


\section{Network morphology and topological study}
\label{subsec: global supervised learning}
In earlier studies, flow networks underwent training to demonstrate the specific function of flow allostery \cite{rocks2019limits,rocks2017designing} through global supervised learning, involving the minimization of a global cost function. This training was primarily focused on ensuring the networks could exhibit a desired pressure drop across a target edge (or multiple target edges) when a pressure drop is applied across a source edge. However, as discussed earlier, relying solely on the desired pressure drop across a target edge is insufficient for defining the equilibrium state of bistable networks, as it neglects the consideration of binary states. 

In this section, we outline two distinct methodologies for the analysis of bistable flow networks, tailored to their specific requirements. Initially, we explore a scenario where the network operates without external pressure forcing, functioning as a closed physical system. Within this framework, the system is capable of adopting $2^{N}$ different binary confidurations, with the resulting configuration of convergence being influenced by both the total volumetric flow within the network and its topology. We postulate an initial state wherein the nodes are devoid of gauge pressure; subsequently, a volumetric flux is introduced into the network through specific nodes over a predetermined duration. 
The primary challenge lies in identifying topology and a set of viscous resistances that drive the system towards a preferred equilibrium from the potential equilibrium configurations. To address this, we employ a \textit{global supervised learning} approach, which will be discussed later.
In the subsequent analysis, our focus shifts to networks wherein the input nodes are subjected to prescribed boundary conditions pressures. There, the objective transitions to determining a resistance configuration that aligns the pressures and binary states at the output nodes with targeted values. This process is facilitated through the utilization of a \textit{local physical supervised learning} method.

\subsection{Training via supervised global learning}
\label{subsub: Global learning}
Here, we present a methodology for determining a flow network topology and resistance to perform multiple tasks using the same acquired structure. To this end, our approach involves optimizing a global cost function for bistable flow networks.

In this context, we aim to use a training algorithm in order to reach various equilibrium states  based on different sets of sources (out of the $2^{N}$ possible states). More specifically, assuming a series of vectors $\{\mathbf{q}^{(h)}(t)\}_{h=1}^{k}$, for each vector $\mathbf{q}^{(h)}(t)$ the system is expected to converge to a (distinct) final state denoted by $\mathbf{v}_{ss}^{(h)}$. For each vector $\mathbf{q}^{(h)}(t)$, we establish a target volume vector $\mathbf{v}_{t}^{(h)}$. The primary goal is to identify a particular Laplacian matrix $\mathbf{W}$ (which represents the topology and the tubes' resistance) that guarantees the system's convergence to the state closest to the specified target state, for each task.

\begin{figure}[]
    \centering
    \includegraphics[width=1\textwidth]{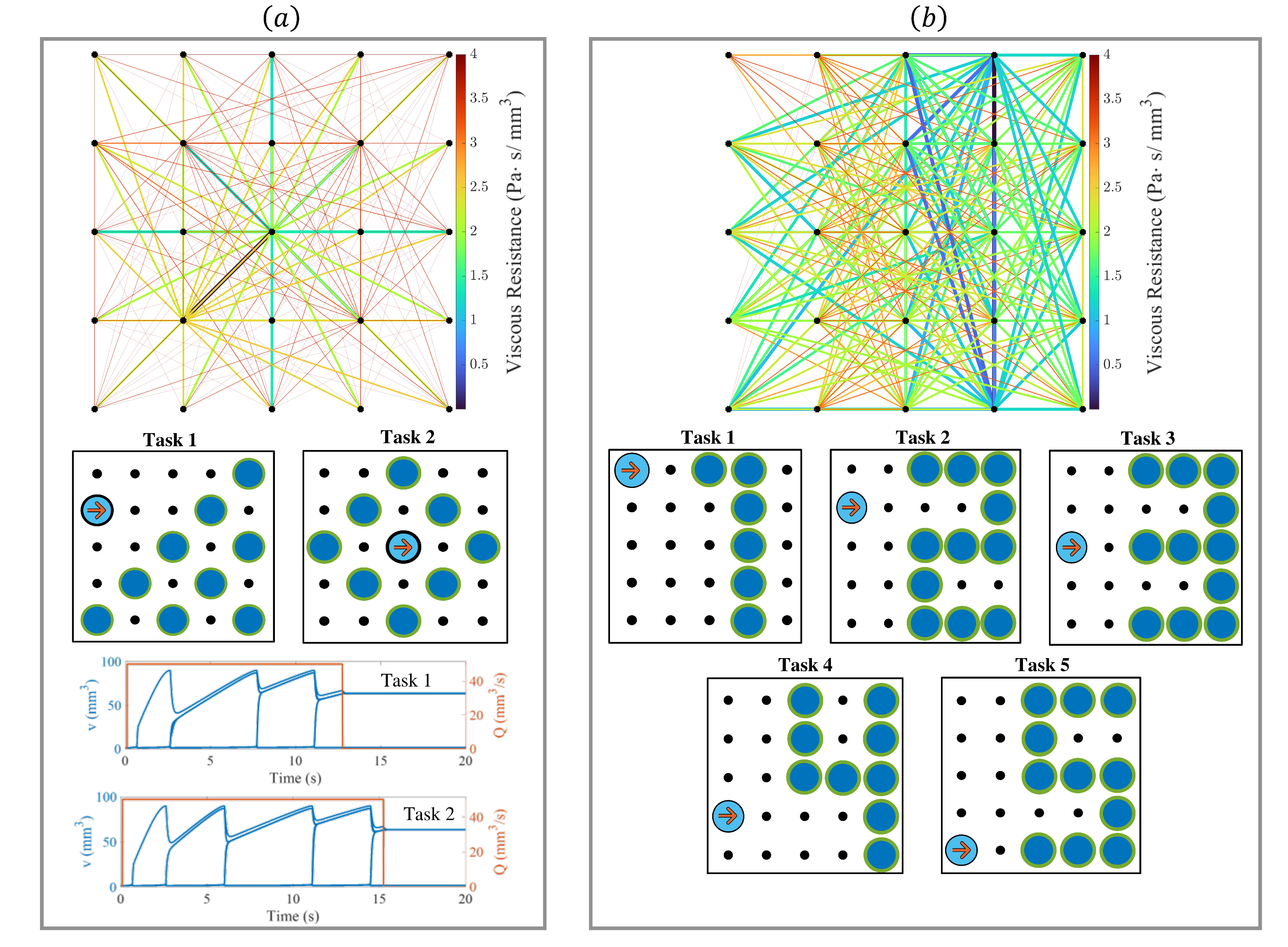}
    \caption{Illustration of two numerical simulations showcasing the learning capabilities of a metamaterial composed of a bistable flow network. The elastic nodes are modeled as hyperelastic balloons, utilizing the Ogden model \cite{Ogden}. These balloons incorporate a bistable regime (between 0.8Pa to 1.1Pa), transitioning from binary state `0' to `1' at a volume of 2.55cc and reverting from state `1' to `0' at 22cc. Learning parameters included learning rate $\eta=0.1$ (see Eq.\,(\ref{eq: PGD})) and $\beta=10^{-5}$. steady state convergence pressure was set at 0.9Pa. A constant flux was introduced into the system during a specific period of time, allowing it to reach equilibrium. The initial viscous resistances were set to unity. Panel (a) displays the results for the first network, indicating the obtained connection topology, colored by viscous resistance and thickness. The middle section depicts the system results in equilibrium, highlighting the inlet and target balloons. The lower part shows volume changes over time (in blue) and the entering flow (in red), obtained by numerical integration of (\ref{eq: Equation of motion}) using ode45 in Matlab. Those nodes that do not snap to the binary state `1' remain nearly at their initial volume, while others snap through in a sequence defined by the resistance configuration. Some snapped nodes exhibit so similar dynamic responses that their graphs appear virtually indistinguishable. Panel (b) exhibits the results for the second network, which mastered five tasks, visually displaying the digit representing the inserted flux's entry balloon in the input layer.}
    \label{fig:Global Metamaterial}
\end{figure}

The volume state vector  of the flow can be (implicitly) calculated through the time integration of the equation of motion (\ref{eq: Equation of motion}), i.e.,
\begin{equation}
\label{eq: v from integraion}
    \mathbf{v}^{(h)}(t)=\mathbf{v}_{0}+\int_{0}^{t}{\mathbf{q}^{(h)}(\tau)\mathrm{d}\tau}-\mathbf{W}\int_{0}^{t}{\mathbf{p}\big(\mathbf{v}^{(h)}(\tau)\big)\mathrm{d}\tau}.
\end{equation}
Upon the system achieving a steady state (or an equilibrium), the steady state volume vector is calculated by $\mathbf{v}^{(h)}_{ss}=\lim_{t\rightarrow\infty}{\mathbf{v}^{(h)}(t)}$.
We note that for any Laplacian matrix $\mathbf{W}$, the relation of the total volume (\ref{eq: v}) holds. This can be easily demonstrated by projecting the vector (\ref{eq: v from integraion}) onto $\mathbf{1}_{N}$ and utilizing the fact that the sum of the columns of $\mathbf{W}$ equals zero. The requirement for the $h^{th}$ target vector is that $\mathbb{V}^{(h)}_{ss}=\mathbf{1}_{N}\cdot\mathbf{v}_{t}^{(h)}$, ensuring compliance with the overall volume constraint. 
The \textit{loss-function} is defined as the Euclidean norm of the volume state vector and the corresponding target vector in the steady state, namely
\begin{equation}
\label{eq: loss-function}
    \mathcal{L}=\cfrac{1}{k}\sum_{h=1}^{k}\big|\big|\mathbf{v}^{(h)}_{ss}-\mathbf{v}^{(h)}_{t}\big|\big|_{2}^{2}.
\end{equation}
To determine a specific Laplacian matrix that ensures the system's convergence to the target state, we introduce the following optimization problem:

\begin{alignat}{2}
\begin{split}
\label{eq: optimization}
 \underset{\mathbf{W}}{\text{minimize}}\,\,\,\,\,\,\,\,\,\,\,\,
 &\mathcal{L}(\mathbf{W})+\beta\big|\big|\mathbf{W}\big|\big|_{2}^{2} \\
 \text{subject to} \quad
 &W_{ij} = W_{ji} \leq 0, \quad i\neq j\\
 &\mathbf{W}\mathbf{1}_{N}=\mathbf{0}_{N},
\end{split}
\end{alignat}
where $\mathbf{0}_{N}$ denote the $N$-vector of all zeros entries and the positive scalars $\beta$ is a regularization parameter. The constraints ensure that the learned Laplacian matrix ($\mathbf{W}$) is valid and positive semidefinite.  
A closed-form solution to the problem appears unattainable, primarily due to the non-convex nature of the problem and the absence of an explicit presentation of the dynamic solution. To address this, we have employed the \textit{projected gradient descent} algorithm (PGD) for a numerical approach to solving the optimization problem \cite{jain2014iterative}. For more details, the reader is encouraged to refer to appendix \ref{Ax: Loss function}. 
Moreover, the minimization problem outlined in (\ref{eq: optimization}) is addressed in appendix \ref{algorithms} by Algorithm \ref{alg:Global}.
The subsequent part showcases various numerical examples of networks where the learning algorithm has successfully converged.

\subsubsection{Numerical demonstrations}
We highlight the successful training algorithm of three bistable networks using the methodology described above. 
In Fig.\,\ref{fig:Global Metamaterial}, we present two numerical simulations demonstrating the learning capabilities inherent in a bistable flow network. The elastic nodes are implemented as hyperelastic balloons, parameterized by the Ogden model \cite{Ogden}, which exhibit a bistable pressure-volume relation. Key learning parameters, including learning rate factor $\eta=0.1$ (see Eq.\,(\ref{eq: PGD})) and $\beta=10^{-5}$, were incorporated. The system underwent a controlled introduction of constant flux within specific time intervals to facilitate the attainment of equilibrium.

The first network contains a 5x5 lattice, consisting of 25 nodes. This lattice was trained with a dual-task objective: a distinct set of outlet nodes experienced a binary state transition, resulting in the visual appearance of diagonal lines below the input node. Meanwhile, the binary states of the remaining internal nodes were set to `0'. In the second task, when a single inlet node transitioned to binary state `1', a specific set of target nodes also underwent a binary state transition to `1', resulting in the formation of a visual rhombus around the inlet node. Once more, the internal nodes maintained a state of `0' (see Fig.\,\ref{fig:Global Metamaterial}(a)).
The second successfully trained network, also comprising a 5x5 lattice of nodes, identifies the left column as an input column. Each node in this column is assigned a specific input number, from 1 at the top to 5 at the bottom. By introducing an influx in the input column and transitioning one of the nodes to binary state `1', the network's task is to register the digit of the nodes where the flow is inserted. The process involves activating the corresponding nodes to the binary state `1' while keeping the other nodes in a state of '0' (see Fig.\,\ref{fig:Global Metamaterial}(b)). 

The obtained outcomes showcase a pioneering concept that can be leveraged in soft robotics actuation and microfluidics applications, particularly in scenarios involving pressurized flow within cavities embedded in elastic bodies. The conventional approach to achieving complex deformation patterns involves the intricate control of multiple inputs, adding complexity to the system's operation \citep{overvelde2015amplifying,che2018viscoelastic,tang2020leveraging,gorissen2020inflatable,wang2023untethered,kaarthik2022motorized,zolfagharian2020closed,ma2019origami,wu2018dial,kaufmann2022harnessing}. However, in our work, we introduce a novel result demonstrating single-input control over a lattice composed of bistable elastic chambers interconnected by thin tubes. This innovative technique holds considerable promise for the development of sophisticated soft actuators with streamlined and extremely efficient control mechanisms \cite{gorissen2019hardware,melancon2022inflatable,che2018viscoelastic,ben2020single,novelino2020untethered}.

Each of the bistable nodes in the network can be viewed as a binary digit. It follows that the entire multistable PNN can be viewed as a memory storage device. The same applies to any multistable system. The critical challenge in such multistable networks is the ability to generate and store information at certain locations in the network, to access that information, and to use that information during computations. Consequently, our algorithm uses two signals spaced apart by a duration exceeding the time required for the network to stabilize, so that after the first signal has been received, the network will have reached a stable equilibrium state, which will act as a memory.

\begin{figure}[]
    \centering
    \includegraphics[width=0.95\textwidth]{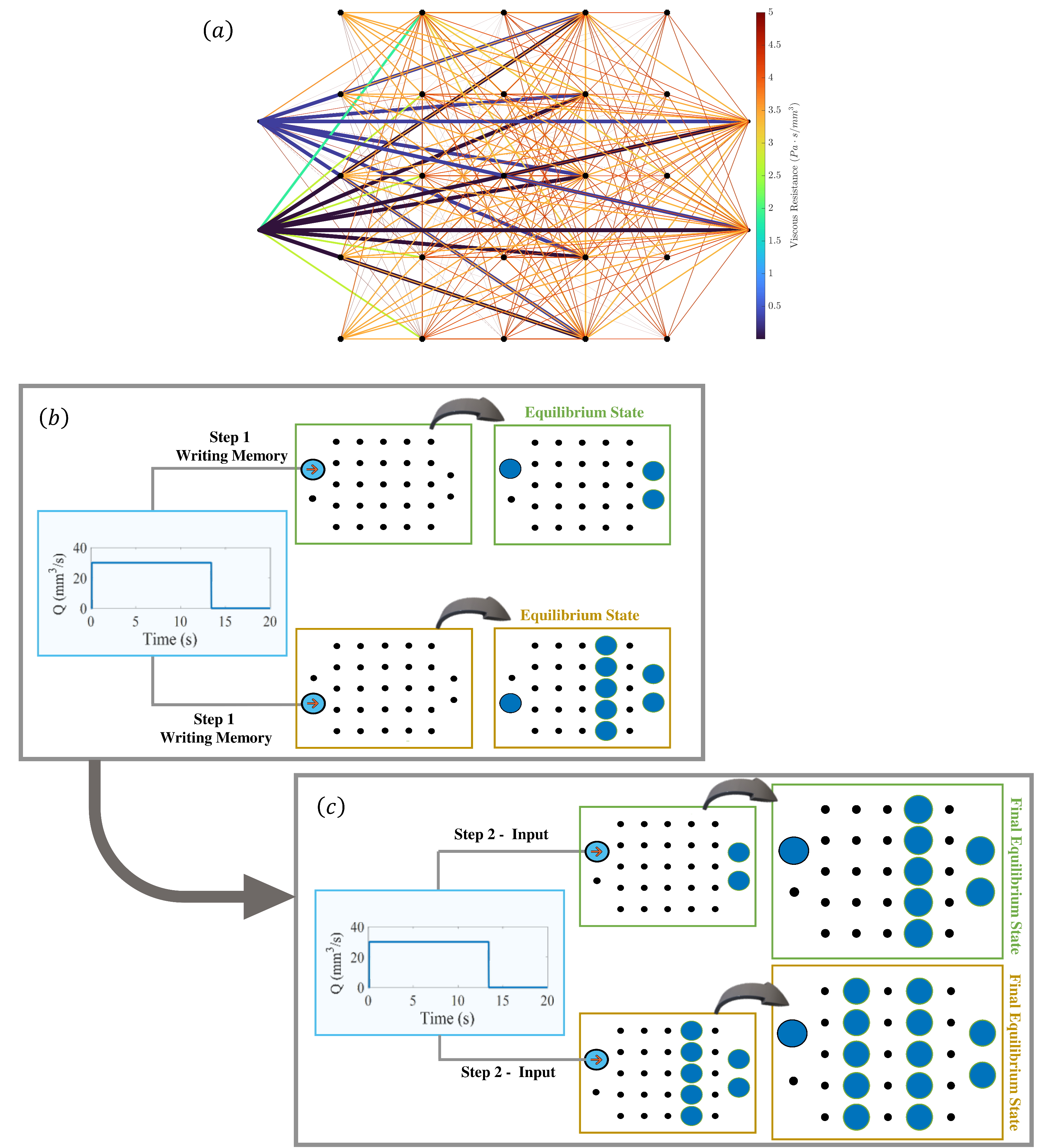}
    \caption{Demonstration of bistable PNNs memory. The input layer of this network is composed of two nodes. The next layer is composed of a 5x5 node lattice, with two nodes as the final layer. We examine two identical topology structures (a), highlighted with green and orange frames. Flow signals are directed to nodes 1 and 2 in each structure's input layer. Once both networks have reached equilibrium (b), both are given the same input signal through node number 1, and are allowed to stabilize again (c). There is one column of nodes in the binary state `1' within the lattice of the first network, and all other nodes are in the binary state `0'. In contrast, the equilibrium state of the second network features two columns of nodes in the binary state `1,' which indicates that the network's response can vary significantly depending on history and, therefore, the initial state. }
    \label{fig:Global MetamaterialMemory}
\end{figure}

In Fig.\,\ref{fig:Global MetamaterialMemory} we demonstrate the results of this approach, by utilizing memory during computation for multistable PNNs that have been trained using the above-described global learning algorithm. Two nodes form the input layer of this network, which is followed by a 5x5 node lattice with two nodes forming the final layer of the network. Two identical PNNs are presented, highlighted with green and orange frames in Fig.\,\ref{fig:Global MetamaterialMemory}(b-c). Each of the identical PNNs receives a different initial input flow signal. The inputs are identical in magnitude and duration, but directed to different nodes, specifically to numbers 1 and 2 in the input layer of each structure. As soon as the systems have stabilized and reached equilibrium, they are then disconnected from external loads, retaining the embedded memory of the initial input in column 4 (see Fig.\,\ref{fig:Global MetamaterialMemory}(b)).
Following this step, both networks receive the same secondary input signal through node number 1, and are once again allowed to stabilize (Fig.\,\ref{fig:Global MetamaterialMemory}(c)). Within the lattice, the equilibrium state of the first network reveals a single column of nodes in the binary state `1', with all other nodes in the binary state `0'. In contrast, the equilibrium state of the second network features two columns of nodes in the binary state `1', demonstrating that the network's response is affected by both the current, and previous inputs. Thus, we show that the computation uses the memory stored during the initial input, to decide on the mechanical output for the secondary input. 

\subsection{Training via local physical-supervised learning}
\label{subsubsec: local supervised learning}

Transport of materials in both biological networks, such as vascular systems, and engineered counterparts, like microfluidic networks, commonly rely on fluid flow. The pipes' properties, such as radii, conductance, and capacitance, collectively influence the network's global material transport capabilities, making it effective. Although computational optimization is a viable strategy, many natural systems tend to adjust individual elements based on localized feedback. In some instances, the flow within pipes can interact with the mechanical characteristics of these tubes, leading to localized adaptations such as constriction or expansion. These adaptations serve to regulate local flow conductance and pipe capacitance, providing the network with a mechanism to control cargo distribution. For example, in \textit{Physarum polycephalum}, the thickness of tubes governs the organism's shape, enabling it to move, forage, and memorize features of its environment \cite{marbach2021network,kramar2021encoding,tero2010rules}. Similarly, adaptive processes in other natural flow networks, such as those in leaves and vasculature \cite{katifori2010damage,ronellenfitsch2016global}, operate through local rules, without a centralized controller. Theoretical explorations of such flow networks have inquired into their potential to learn diverse behaviors through local rules, including classifying stimuli reminiscent of machine learning principles \cite{stern2021supervised,anisetti2023learning,scellier2021deep,dillavou2022demonstration}.

Here, we focus on bistable networks with sustained pressure applied to their external nodes (as a boundary condition). These systems always reach a stable steady state, requiring a period of relaxation. 
The main objective is to align the steady state output nodes' binary states (and corresponding pressures) with the predefined targets. 
During supervised learning, inputs and outputs are specified. To achieve these desired output pressures, the learning degrees of freedom of some tubes must be fine-tuned.
The algorithm designed for instructing the system to attain the prescribed pressure has been described in the works of \citet{dillavou2022demonstration} and \citet{stern2021supervised} for other applications under the assumption that networks maintain linear relationships. Their algorithm is briefly described in the appendix.\,\ref{Local_algo_old}. In the context of a bistable network, however, it becomes imperative to modify the algorithm steps. 
In the following discussion, key adjustments to the algorithm are outlined. 

Since the relation between pressure and node volume are complex, nonlinear, and bistable, our initial emphasis is on making sure each node's binary state aligns with the target binary state, representing the desired result. Subsequently, careful fine-tuning of the viscous resistances ensues to guide the nodes to the required pressure within the correct binary state.
This algorithm hinges on the memory property inherent in bistable networks. The system converges to a stable steady state at every stage. This stable state serves as the initial condition for the subsequent iteration of the learning. Through such a mechanism, the system can attain unique convergence basins that would not be possible without preserving intermediate states.
The pseudo-code for our learning algorithm can be found in Algorithm \ref{alg:Local} in appendix \ref{algorithms}.

Throughout the training regimen, changes to viscous resistances are executed through a comparative analysis of two distinct flux states imposed on the same network by different boundary conditions - denoted as free and clamped networks. In the free state, the network attempts to accomplish the designated task by applying input pressures, $\mathbf{p_{BC}}$, and subsequently producing corresponding output pressures, $\mathbf{p_{out}^{f}}$, and binary states for each output node. In the clamped state, identical inputs are applied, $\mathbf{p_{BC}}$, but additional pressures are applied at the output nodes. In this context, we distinguish two scenarios: when the binary state of the node is not acceptable (i.e., does not meet the target binary state), this node shall transition between the binary states. Specifically, when transitioning the node from binary state `0' to binary state `1', the dictated pressure exceeds the local maximum point in the pressure-volume characteristic. Conversely, when transitioning the node from binary state `1' to `0', the dictated pressure is set below the local minimum point in the pressure-volume characteristic. 
In cases where the binary state of the output node is acceptable
(i.e., meeting the target), the dictated pressure follows the \citet{dillavou2022demonstration} and \citet{stern2021supervised} algorithm as described in the appendix.\,\ref{Local_algo_old}.
In addition, implementing our local learning algorithm requires numerically solving $N$ nonlinear ODEs during each iteration. Those networks with extensive or significant resistance can have inefficient, inefficient, or non-convergent computational demands. In appendix\,\ref{Algebraic algorithm} we describe our alternative algorithm for optimizing the training methodology without repeating differential equation resolutions.

\subsubsection{Numerical demonstrations}
\begin{figure}
  \centering 
  \includegraphics[width=0.9\textwidth]{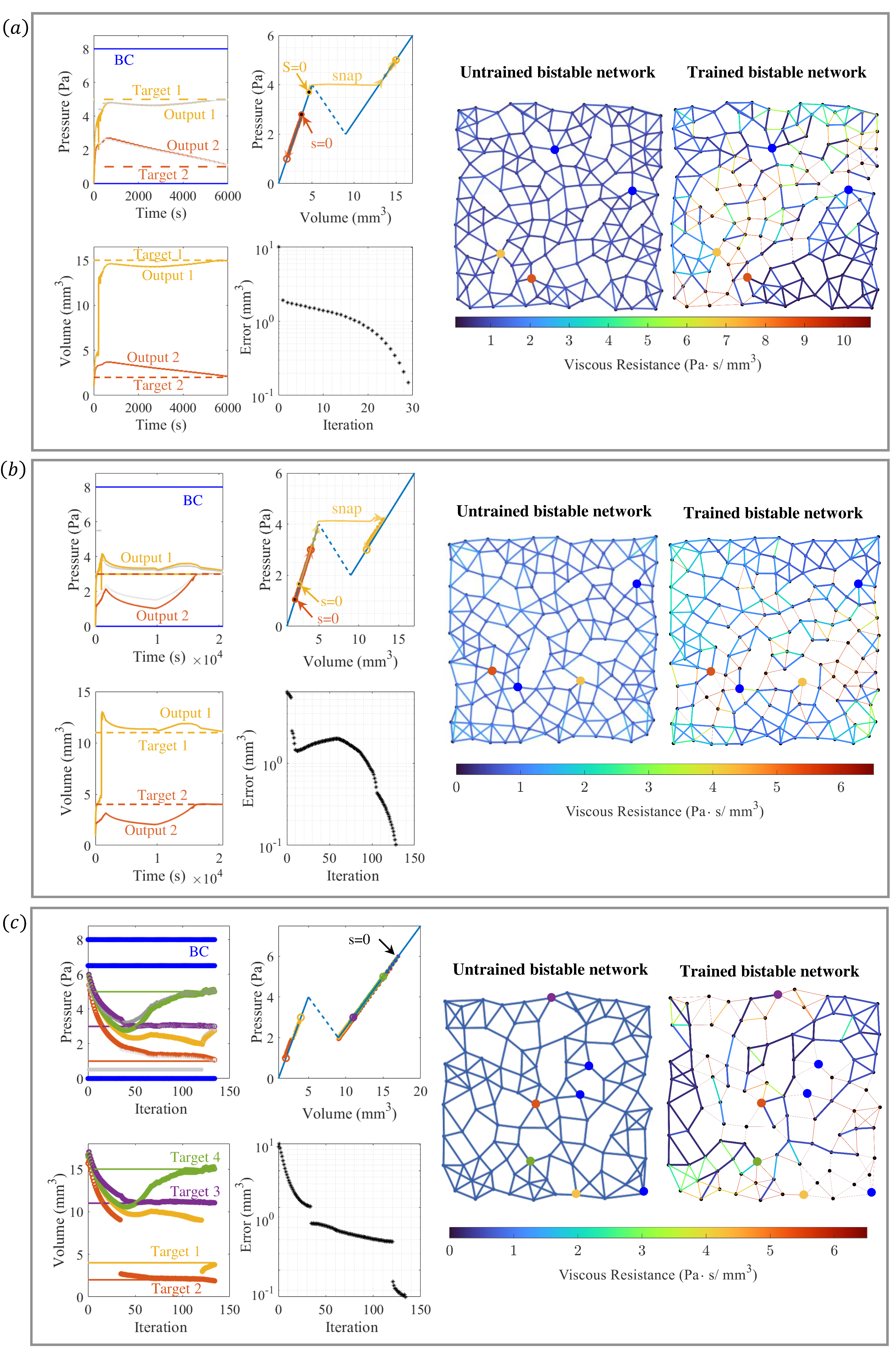}
  \caption{(Caption next page.)}
  \label{fig:local Metamaterial}
\end{figure}

Below, we present three simulations showcasing the success of the learning process for bistable flow networks utilizing the aforementioned algorithm.
Initially, we devised a disordered network of nodes, randomly located in the plane, stipulating that the minimum distance between adjacent nodes exceeds a predefined threshold. The nodes were interconnected by tubes, ensuring that each node was linked to up to the five closest nodes, under the condition that the distance did not surpass a predefined value.
The (initial) viscous resistances of the network were defined proportionally to the distance between nodes. We began with a volume of 1cc, meaning that the gauge pressure was zero. In Fig.\,\ref{fig:local Metamaterial}, we illustrate training outcomes for two networks comprising 150 nodes each and another network with 100 nodes. In each simulation, the input nodes were randomly designated (highlighted in blue), and output nodes were also randomly selected (highlighted in orange and yellow for two exit nodes, or yellow, orange, purple, and green for four exit nodes). 
In the presented simulations, we employed a theoretical bistable characteristic that correlates pressure to the volume of nodes, approximating a simple trilinear curve for simulation simplicity. This characteristic delineates two distinct binary states: state `0' for volumes smaller than 5cc and state `1' for volumes greater than 9cc (recall that the middle branch, the spinodal, is unstable and is denoted by a dashed line). The displayed error, labeled as \textit{Error}, is defined by the squared norm difference between the volumes obtained in the simulation and the volumes defined as the target. We examine the difference between volumes rather than pressures since the nodes are characterized by a trilinear curve, and tracking pressure does not uniquely describe the network's state.

\addtocounter{figure}{-1}
\begin{figure}
  \caption{Illustration of three numerical simulations demonstrating the local-learning algorithm of a bistable flow network. For the simulation, we set the parameters $\eta=0.25$, $\gamma=0.01$, and defined the convergence when the error reaches $10^{-1}$. Theoretical bistable characteristics, simplifying to a trilinear curve, were employed for simulation. The unstable spinodal is denoted by a dashed line. The \textit{Error} indicates the squared norm difference between simulated and target volumes. The first learning iteration is indicated by the notation $s=0$. Panels (a)-(b) illustrate training outcomes for two 150-node networks, in which blue nodes represent input nodes, and orange/yellow output nodes aim to reach specified pressures and binary states. Gray lines represent supervisor pressure constraints. A series of panels shows the training process through pressure-time, volume-time, pressure-volume, and error iterations. Visualization includes initial and final networks, with tube colors and thickness indicating viscous resistance. In panel (c), results of a 100-node network achieving four targets are presented, with each point representing the converged steady state pressure or volume.}
\end{figure}

In the initial simulation, presented in Fig.\,\ref{fig:local Metamaterial}(a), a network comprising 150 nodes was defined. Two nodes were set to pressures of 0Pa and 8Pa, with two specific goals outlined: one node aimed to achieve a pressure of 5Pa and be in a binary state `1', while another node was to reach a pressure of 1Pa and be in a binary state `0'. The algorithms of \citet{dillavou2022demonstration} and \citet{stern2021supervised} could be employed for training in this scenario; however, the enhanced algorithm we introduced significantly optimizes the learning process and reduces the number of iterations needed for convergence.
In the initial iteration, the first output node (marked in yellow) converged to an incorrect binary state regarding the target. However, in the subsequent iteration, owing to our learning process, the node converged to the desired state. Consequently, the error is also discontinuously decreased, as expected during the "snap-through" transition of the system through an unstable branch.
Fig.\,\ref{fig:local Metamaterial}(a)-(b) displays multiple panels, including pressure as a function of time representing the pressures of the output nodes. Other panels illustrate volume as a function of time, pressure as a function of volume, and the error as a function of iterations. Additionally, visualizations of the initial (untrained) network and the final trained network are presented. The colors of the tubes are mapped according to the scale of the viscous resistances corresponding to those tubes, and the thickness is proportional to the viscous resistance (i.e., greater resistance corresponds to smaller line thickness, and vice versa). Remarkably, the network demonstrated successful training after only 30 iterations in this simulation.

In the second simulation, we modified the target pressures to lie within the bistable region of the pressure-volume curve, setting them at 3Pa. For one output node, we specified it to be in binary state `0', while for the second output node, it was set to be in binary state `1'. Similar to the preceding simulation, the success of the learning algorithm is evident, with the network converging after only 128 iterations. Based on this simulation, the learning process presented in this paper has several strengths.
Firstly, the capability of the first output node to reach the binary state `1' and reside within the bistable region is achieved exclusively through the system's memory feature. This highlights the indispensability of this feature for attaining the requisite state. Additionally, the algorithm successfully discriminates between the distinct binary states of the nodes, even when the target pressures are identical.

In the third simulation, we present the training results of a 100-node network tasked with achieving four goals. The objectives incorporate elements from the two prior experiments, requiring two output nodes to reach pressures of 1Pa and 5Pa, and an additional two output nodes to attain identical pressures of 3Pa but at different binary states. Furthermore, three input nodes were defined in this simulation with forced pressures of 7Pa and 8Pa.
In Fig.\,\ref{fig:local Metamaterial}(c), we depict slightly different panels than before, showcasing pressure and volume as functions of the iteration number. Each point in the graph represents the pressure (or volume) to which the system has converged in a steady state. Specifically, in the first iteration, the four outlet nodes reached pressures higher than required. Consequently, during the learning process, two abrupt transitions were executed to attain appropriate binary states (from `0' to `1'), demonstrated by the sharp jumps in the volume graph for output nodes 1 and 2 and the Error graph. This simulation also demonstrates excellent convergence of the learning algorithm, indicating success in training a bistable flow network to accomplish multiple tasks.

\begin{figure}[h]
    \centering
    \includegraphics[width=1\textwidth]{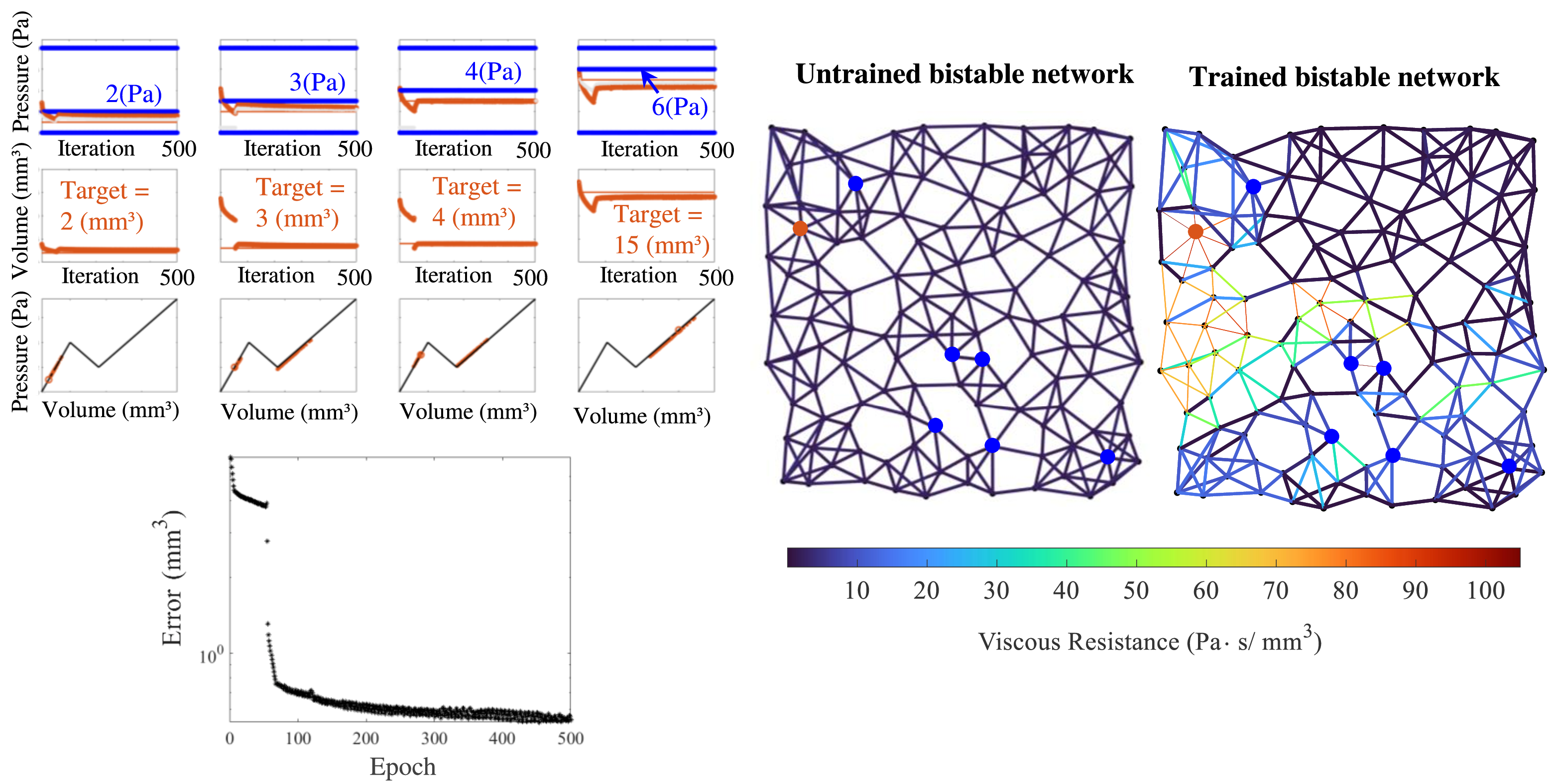}
    \caption{Numerical simulation conducted on a grounded network with 150 nodes and five inputs, aimed at training the output node for multiple tasks with varying inputs. The elastic characteristic (pressure-volume relationship) of each node corresponds to the profile depicted in Fig.\,\ref{fig:local Metamaterial}. Parameters were set as $\eta=0.25$, $\gamma=0.01$, with convergence defined by an error threshold of $5\times 10^{-1}$, where "Error" denotes the mean of squared norm difference between simulated and target volumes for each task. Constant inlet pressure was maintained at 8Pa (and 0Pa), while other inlets varied between 2Pa, 3Pa, 4Pa, and 6Pa, targeting output volumes of 2cc$^3$, 3cc$^3$, 4cc$^3$, and 15cc$^3$, respectively. Blue nodes indicate inputs, with the orange output node reflecting the aim to achieve specified pressures and binary states. Both initial and final network states with variations in vortex colors and thickness illustrate changes in viscous resistances.}
    \label{fig:local Metamaterial22}
\end{figure}

Finally, Fig.\,\ref{fig:local Metamaterial22} showcases the outcomes of a simulation conducted on a grounded network comprising 150 nodes with five inputs. This simulation aims to train the output node to execute multiple tasks under varying inputs. Accordingly, one inlet pressure was maintained constant across all tasks at 8Pa, while the pressures at the other four inlets were adjusted task-specifically to 2Pa, 3Pa, 4Pa, and 6Pa. Correspondingly, the targets for the output node's volume were set at 2cc$^3$, 3cc$^3$, 4cc$^3$, and 15cc$^3$ for each input pressure.
An epoch in this context was structured around four learning iterations, with the network undergoing simultaneous training across all tasks. The modification in resistances required for accomplishing each task was computed, and the adjustment implemented in each epoch represented the average of these calculated modifications. The simulation evidenced convergence after 500 epochs. 

\section{Concluding remarks}
\label{sec:Results}

In this work, we combined multistability with the concept of physical neural networks and studied the properties of such physical systems. The discussion below summarizes the current work and presents the broader implications of our findings. 

Our focus was on bistable flow networks. Flow networks are characterized by internal, external, and output nodes, facilitated by either constant pressures or volumetric flow rates. The bistable nature of each network element, presenting a non-linear pressure-volume relationship and two distinct states of positive stiffness (`0' and `1'), enables the versatility and adaptability of this system. We examined the network's potential equilibrium configurations or steady states, analyzed their stability, and harnessed the complex dynamic behavior of the system. Addressing these aims necessitated the development of two methodologies, each customized to suit the network's specific operational demands. 

The analysis of the network's steady states and the stability of its states highlights that, in scenarios devoid of external flux or prescribed pressure, the network will reach a state of pressure equilibrium regardless of its topological configuration. Conversely, when external pressures are introduced, the resultant steady state pressures are determined by the interplay between the network's topology and the resistances within its connecting tubes. Although the existence of steady state configurations may be contingent upon these parameters, the stability of the network at a steady state was found to be invariant with respect to both topology and resistance values. This insight underscores the robustness of such networks against structural and operational variabilities. 

Having studied the equilibrium dynamics and stability of such networks, we investigated their complex dynamics under the control of inlet volumetric flux, which was specified for a limited duration, and revealed the network's ability to adopt distinct binary states that are influenced by its topological structure. This stage emphasized the challenge of identifying a configuration of network topologies and resistances to guide the system toward a desired equilibrium state. A global supervised learning algorithm was used to address this challenge. These results provided a novel concept with significant applications for the fields of soft robotics and microfluidics, especially in contexts that necessitate pressurized flow within cavities of elastic materials. Traditionally, generating complex deformation patterns in such systems requires the precise manipulation of numerous inputs, complicating their operation. Thus, a significant interest exists in simplifying the control of such systems \cite{polygerinos2017soft,marchese2015recipe,unger2000monolithic,thorsen2002microfluidic,desai2012design,mosadegh2010integrated}, and our research presents a new strategy for designing bistable elastic chamber networks controlled by a single input. Soft actuators can be improved by using this approach, which offers a more streamlined and efficient system for controlling them, potentially enhancing their design and functionality. As a result, more advanced soft robotic systems and microfluidic devices may be developed, reducing operational complexity while enhancing performance and control accuracy.

Finally, we applied a local supervised learning method. This learning strategy, drawing upon the mechanical attributes of the network's elements, enables the network to progress towards defined operational objectives. This part of our study reveals the potential of these networks to perform an array of tasks—such as interpolations, regressions, sorting, and classification—using purely mechanical networks. 
This study also raised some limitations which are discussed in Appendix\,\ref{Limitations}.

We have explored and demonstrated the potential of bistable physical neural networks, emphasizing their ability to perform a diverse array of tasks, purely mechanically. This illustrates a paradigm where mechanical systems mimic the versatile, task-specific functionality characteristic of digital neural networks in the realm of artificial intelligence.
This research highlights the network's ability to function as a memory storage system. Each bistable node within the network operates as an independent data storage unit. Consequently, every stable equilibrium configuration effectively acts as a `stored memory', allowing the network to maintain the last reached state, even when disconnected from any external inputs. This characteristic facilitates the pre-setting of the network to diverse equilibrium states under uniform inputs by pre-encoding memories. This capability not only enhances the adaptability of the network but also broadens its applicability across various computational scenarios.
We demonstrated how the network can write data to a memory storage. By utilizing this property, bistable PNNs can perform computations that involve storing information as memory and use it in subsequent computations.

Our work offers real-life opportunities for innovation in smart technologies, medical devices, and other fields.
Furthermore, our research lays foundational stones towards the realization of computational matter, marking the advent of learning matter. The findings of this study mark an important milestone, signaling new opportunities for future investigation.

\newpage

\appendix

\section{Proofs of propositions regarding the stability analysis} 
\label{Proofs}
This appendix provides mathematical proofs for the propositions asserted in the stability analysis in section \ref{subsection: Network equilibrium stability}. 
\subsection{Proof of proposition. \ref{prob:Reduced EOM}}
The flow balance on the $i^{th}$ node yields: 
\begin{equation}
\label{eq: Ap1-flow balance}
    \frac{\mathrm{d}v_{i}}{\mathrm{d}t}=q_{i}+\sum_{\substack{j=1 \\ j \neq i}}^{N}{C_{ij}\big(p_j(v_j)-p_i(v_i)\big)};\qquad i=1,\cdots,N.
\end{equation}
Since we realized that
\begin{equation}
\label{eq: Ap1-p-psi}
    p_{i}(v_{i})=\upsilon\frac{\mathrm{d}\psi}{\mathrm{d}v_{i}},
\end{equation}
substitution in Eq.\,(\ref{eq: Ap1-flow balance}) lead to the governed equation of the $i^{th}$ node, 
\begin{equation}
\label{eq: Ap1-flow balance index}
    \frac{\mathrm{d}v_{i}}{\mathrm{d}t}=q_{i}+\upsilon\sum_{\substack{j=1 \\ j \neq i}}^{N}{C_{ij}\bigg(\frac{\mathrm{d}\psi}{\mathrm{d}v_{j}}-\frac{\mathrm{d}\psi}{\mathrm{d}v_{i}}\bigg)}=q_{i}+\bigg(\upsilon\sum_{\substack{j=1 \\ j \neq i}}^{N}{C_{ij}\frac{\mathrm{d}\psi}{\mathrm{d}v_{j}}}\bigg)-\upsilon\frac{\mathrm{d}\psi}{\mathrm{d}v_{i}}\sum_{\substack{j=1 \\ j \neq i}}^{N}{C_{ij}}.
\end{equation}
By eliminating the equation of motion for the $N^{th}$ node in Eq.\,(\ref{eq: Ap1-flow balance}) under the known volume constraints, $v_{N}=\mathbb{V}_{SS}-(v_{1}+\cdots+v_{N-1})$, and employing the chain rule, the pressure $p_{i}(v_{i})$ mentioned in Eq.\,(\ref{eq: Ap1-p-psi}) can be expressed as:
\begin{equation}
\label{eq: constrained derivative}
    p_{i}(v_{i})=\upsilon\bigg(\frac{\mathrm{d}\psi}{\mathrm{d}v_{i}}+\frac{\mathrm{d}\psi}{\mathrm{d}v_{N}}\frac{\mathrm{d}v_{N}}{\mathrm{d}v_{i}}\bigg)=\upsilon\bigg(\frac{\mathrm{d}\psi}{\mathrm{d}v_{i}}-\frac{\mathrm{d}\psi}{\mathrm{d}v_{N}}\bigg);\qquad i=1,\cdots,N-1.
\end{equation}
By substituting Eq.\,(\ref{eq: constrained derivative}) into Eq.\,(\ref{eq: Ap1-flow balance}) and considering only the $N-1$ elements (excluding the $N^{th}$ element), the result aligns with Eq.\,(\ref{eq: Ap1-flow balance index}) without reference to the $N^{th}$ node. Therefore, for the constrained system, it is feasible to take the general equation of motion as shown in Eq.\,(\ref{eq: Equation of motion}) in its matrix form, eliminate the $N^{th}$ row and column from $\mathbf{W}$, remove the $N^{th}$ element in the vector $\mathbf{q}$, and derive the reduced system
\begin{equation}
\label{eq: constrained equation of motion Appx}
    \frac{\mathrm{d}\tilde{\mathbf{v}}}{\mathrm{dt}}=-\mathbf{\tilde{W}\tilde{p}}+\mathbf{\tilde{q}}=-\mathbf{\tilde{W}\nabla_{\mathbf{\tilde{v}}}^{T}\tilde{\psi}(\mathbf{\tilde{v}})}+\mathbf{\tilde{q}},
\end{equation}
where $\tilde{\mathbf{v}}=[v_{1},\cdots,v_{N-1}]^{T}$ is a reduced state vector.  \(\blacksquare\)

\subsection{Proof of proposition. \ref{prob:Reduced W}}

Firstly, it is clear that given the Laplace matrix is square, real, and symmetric - the removal of its $N^{th}$ row and column results in a matrix that retains these characteristics, thus remaining square, real, and symmetric.
Hence, we aim to demonstrate that the matrix \(\mathbf{\tilde{W}}\), obtained after removing the $N^{th}$ row and column, is positive definite. Consider an arbitrary vector \(\mathbf{u}\), and define the quadratic form \(E(\mathbf{u}) = \frac{1}{2}\mathbf{u}^{T}\mathbf{\tilde{W}}\mathbf{u}\). Our objective is to establish the positivity of this quadratic form for any chosen vector \(\mathbf{u}\). In index notation, the function \(E\) can be expressed as 
\begin{equation}
\label{eq: index notation of E}
    E(\mathbf{u}) = \frac{1}{2}\mathbf{u}^{T}\mathbf{\tilde{W}}\mathbf{u}=\frac{1}{2}\sum_{i=1}^{N-1}\sum_{j=1}^{N-1}{\tilde{W}_{ij}u_{i}u_{j}}.
\end{equation}
Decomposing this into the diagonal terms and the non-diagonal terms (of $\mathbf{\tilde{W}}$), and introducing the general term \(W_{ij}\) into the sums, we arrive at 
\begin{equation}
\label{eq: index notation of E 2}
    E(\mathbf{u}) = \frac{1}{2}\sum_{i=1}^{N-1}{\bigg(C_{iN}u_{i}u_{i}+\sum_{\substack{j=1 \\ j \neq i}}^{N-1}{C_{ij}u_{i}u_{i}}-\sum_{\substack{j=1 \\ j \neq i}}^{N-1}{C_{ij}u_{i}u_{j}}\bigg)}.
\end{equation}
Leveraging the symmetry of the matrix (\(C_{ij} = C_{ji}\)), we further rearrange 
\begin{equation}
\label{eq: index notation of E 3}
    E(\mathbf{u}) = \frac{1}{2}\sum_{i=1}^{N-1}{\bigg(C_{iN}u_{i}u_{i}+\cfrac{1}{2}\bigg(\sum_{\substack{j=1 \\ j \neq i}}^{N-1}{C_{ji}u_{i}^{2}}+\sum_{\substack{j=1 \\ j \neq i}}^{N-1}{C_{ij}u_{j}^{2}}\bigg)-\sum_{\substack{j=1 \\ j \neq i}}^{N-1}{C_{ij}u_{i}u_{j}}\bigg)},
\end{equation}
and thus, we obtain 
\begin{equation}
\label{eq: index notation of E 4}
    E(\mathbf{u}) = \frac{1}{2}\sum_{i=1}^{N-1}{C_{iN}u_{i}^{2}}+\frac{1}{2}\sum_{i=1}^{N-1}\sum_{\substack{j=1 \\ j \neq i}}^{N-1}{\frac{1}{2}C_{ij}(u_{i}-u_{j})^{2}}>0.
\end{equation}
The resulting sum comprises positive terms. 
Consequently, the quadratic form we derived is indeed real positive. \(\blacksquare\)

\subsection{Proof of proposition. \ref{prob:Reduced T}}
Having established that \(\tilde{\mathbf{W}}\) is real, symmetric, and positive definite, then there exists an orthonormal matrix \(\mathbf{P}\) such that \(\tilde{\mathbf{W}}=\mathbf{PDP}^{T}\), where \(\mathbf{D}\) is a diagonal matrix with diagonal entries being the eigenvalues of \(\tilde{\mathbf{W}}\), further \(\mathbf{P}^{T}=\mathbf{P}^{-1}\) due to orthonormality, and the column vectors of \(\mathbf{P}\) are linearly independent eigenvectors of \(\tilde{\mathbf{W}}\) that are mutually orthogonal. Since the diagonal elements of \(\mathbf{D}\) are positive, it is given for further decomposition \(\tilde{\mathbf{W}}=\mathbf{\big(PD^{1/2}\big)\big(D^{1/2}P\big)}^{T}\). We denote \(\mathbf{L} = \mathbf{PD^{1/2}}\); hence, it is also an orthonormal matrix since it holds: \(\mathbf{L}^{-1}=\mathbf{D}^{-1/2}\mathbf{P}^{-1}=\mathbf{D}^{-1/2}\mathbf{P}^{T}=\mathbf{L}^{T}\). Consequently, we introduce the matrix \(\mathbf{T} = \mathbf{L}^{-1}\), satisfying 
\begin{equation}
\label{eq: TWT}
  \mathbf{T}\mathbf{\tilde{W}}\mathbf{T}^{T}=\big(\mathbf{L}^{-1}\big)\big(\mathbf{L}\mathbf{L}^{T}\big)\big(\mathbf{L}^{-T}\big)=\big(\mathbf{L}^{-1}\mathbf{L}\big)\big(\mathbf{L}^{T}\mathbf{L}^{-T}\big)=\mathbf{I}.
\end{equation}
\(\blacksquare\)

\subsection{Proof of proposition. \ref{prob:Reduced EOM s}}
Once the invertible transformation matrix $\mathbf{T}\in\mathbb{R}^{(N-1)\times(N-1)}$ is established, we can define a new set of coordinate as $\mathbf{\tilde{s}}=\mathbf{T}\mathbf{\tilde{v}}$. 
We differentiate the relationship with respect to time, and substitute the equation of motion from Eq.\,(\ref{eq: constrained equation of motion}), resulting in: 
\begin{equation}
\label{eq: equation in s}
  \frac{\mathrm{d}\tilde{\mathbf{s}}}{\mathrm{dt}}=\mathbf{T}\frac{\mathrm{d}\tilde{\mathbf{v}}}{\mathrm{dt}}=\mathbf{T}\big(-\mathbf{\tilde{W}}\nabla_{\mathbf{\tilde{v}}}^{T}\mathbf{\tilde{\psi}(\mathbf{\tilde{v}})}+\mathbf{\tilde{q}}\big).
\end{equation}
Following the chain rule, we derive $\nabla_{\mathbf{\tilde{v}}}^{T}\tilde{\psi}(\mathbf{\tilde{v}})=\mathbf{T}^{T}\nabla_{\mathbf{\tilde{s}}}^{T}\psi(\mathbf{\tilde{s}})$. 
We substitute the latter relationship into Eq.\,(\ref{eq: equation in s}) and combine the resulting expression from Eq.\,(\ref{eq: TWT}), yielding:
\begin{equation}
\label{eq: equation in ds}
  \frac{\mathrm{d}\mathbf{\tilde{s}}}{\mathrm{dt}}=-\nabla_{\mathbf{\tilde{s}}}^{T}\tilde{\psi}(\mathbf{\tilde{s}})+\mathbf{\tilde{g}}, 
\end{equation}
where $\tilde{\psi}(\mathbf{\tilde{s}})=\tilde{\psi}(s_{1},\cdots,s_{N-1})$ and $\mathbf{\tilde{g}}=\mathbf{T}\mathbf{\tilde{q}}$.
\(\blacksquare\)

\subsection{Proof of proposition. \ref{prob_effectiv}}
The proposition is straightforward and can be easily concluded through basic differential rules.

\subsection{Proof of proposition. \ref{prob_ss}}
In what follows, we prove that  $\mathrm{d}\mathbf{\tilde{v}}/\mathrm{dt}=\mathbf{0}$ if and only if $\mathrm{d}\mathbf{\tilde{s}}/\mathrm{dt}=\mathbf{0}$.
We employ relation (\ref{eq: equation in s}) and note that the transformation matrix, $\mathbf{T}$, is invertible. Consequently, if $\mathrm{d}\mathbf{\tilde{s}}/\mathrm{dt}=\mathbf{0}$, then $\mathbf{T}\mathrm{d}\mathbf{\tilde{v}}/\mathrm{dt}=\mathbf{0}$, implying only the trivial solution, $\mathrm{d}\mathbf{\tilde{v}}/\mathrm{dt}=\mathbf{0}$. Conversely, if $\mathrm{d}\mathbf{\tilde{v}}/\mathrm{dt}=\mathbf{0}$, it is evident from Eq.\,(\ref{eq: equation in s}) that $\mathrm{d}\mathbf{\tilde{s}}/\mathrm{dt}=\mathbf{0}$. \(\blacksquare\)

\subsection{Proof of proposition. \ref{prob:similarity}}
Our goal here is to establish the similarity between the Hessian matrices of the potential energies $\tilde{\phi}\big(\mathbf{\tilde{s}})$ and $\tilde{\psi}(\mathbf{\tilde{v}})$. 
The Hessian matrix calculated with respect to the constrained state coordinates $\mathbf{\tilde{v}}$ is $\mathbb{H}_{\mathbf{\tilde{v}}}(\tilde{\psi})=\nabla_{\mathbf{\tilde{v}}}^{T}\mathbf{\nabla_{\tilde{v}}\tilde{\psi}}$, and the Hessian matrix of the effective potential energy calculated with respect to $\mathbf{\tilde{s}}$ is $\mathbb{H}_{\mathbf{\tilde{s}}}(\tilde{\phi})=\nabla_{\mathbf{\tilde{s}}}^{T}\mathbf{\nabla_{\tilde{s}}\tilde{\psi}}$. Utilizing the chain rule $\nabla_{\mathbf{\tilde{v}}}^{T}=\mathbf{T}^{T}\nabla_{\mathbf{\tilde{s}}}^{T}$, or equivalently by using the transpose operation $\nabla_{\mathbf{\tilde{v}}}=\nabla_{\mathbf{\tilde{s}}}\mathbf{T}$ , we obtain that $\mathbb{H}_{\mathbf{\tilde{v}}}(\tilde{\psi})=\mathbf{T}^{T}(\nabla_{\mathbf{\tilde{s}}}^{T}\mathbf{\nabla_{\tilde{s}}\tilde{\psi}})\mathbf{T}=\mathbf{T}^{T}\mathbb{H}_{\mathbf{\tilde{s}}}(\tilde{\phi})\mathbf{T}$. 
Leveraging the fact that $\mathbf{T}=\mathbf{L}^{-1}=\mathbf{L}^{T}$ we immediately obtain 
\begin{equation}
\label{eq: similarity}
 \mathbb{H}_{\mathbf{\tilde{v}}}(\tilde{\psi})=\mathbf{L}\mathbb{H}_{\mathbf{\tilde{s}}}(\tilde{\phi})\mathbf{L^{-1}}.
\end{equation}
Thus, the Hessian matrices exhibit similarity. \(\blacksquare\)

\section{Loss function gradient and technical details on the iterative process}
\label{Ax: Loss function}
For completeness, we introduce the loss function gradient tensor with respect to the Laplacian matrix.
Back to our definition of the loss-function in Eq.\,(\ref{eq: loss-function}), Euclidean inner product $\langle \cdot \; , \; \cdot \rangle$ can be used to rewrite the loss function as follows

\begin{equation}
\label{eq: loss function gradient Frobenius}
    \mathcal{L}=\frac{1}{k}\sum_{h=1}^{k}{\left\langle\mathbf{v}_{ss}^{(h)}-\mathbf{v}_{t}^{(h)}\,,\,\mathbf{v}_{ss}^{(h)}-\mathbf{v}_{t}^{(h)}\right\rangle}.
\end{equation}
Therefore, using symmetry known properties of the inner product combined with Eq.\,(\ref{eq: v from integraion}), we can calculate the gradient of $\mathcal{L}$ with respect to $\mathbf{W}$, namely,

\begin{equation}
\begin{aligned}
\label{eq: loss function gradient Frobenius2}
    \frac{\partial\mathcal{L}}{\partial\mathbf{W}}=&\frac{1}{k}\sum_{h=1}^{k}{\bigg[\left\langle\frac{\partial}{\partial\mathbf{W}}\big(\mathbf{v}_{ss}^{(h)}-\mathbf{v}_{t}^{(h)}\big)\,,\,\mathbf{v}_{ss}^{(h)}-\mathbf{v}_{t}^{(h)}\right\rangle+\left\langle\mathbf{v}_{ss}^{(h)}-\mathbf{v}_{t}^{(h)}\,,\,\frac{\partial}{\partial\mathbf{W}}\big(\mathbf{v}_{ss}^{(h)}-\mathbf{v}_{t}^{(h)}\big)\right\rangle\bigg]}=\\
    =&\frac{2}{k}\sum_{h=1}^{k}{\left\langle\mathbf{v}_{ss}^{(h)}-\mathbf{v}_{t}^{(h)}\,,\,\frac{\partial}{\partial\mathbf{W}}\big(\mathbf{v}_{ss}^{(h)}-\mathbf{v}_{t}^{(h)}\big)\right\rangle}=\\
    =&\frac{2}{k}\sum_{h=1}^{k}{\left\langle\mathbf{v}_{ss}^{(h)}-\mathbf{v}_{t}^{(h)}\,,\,\frac{\partial\mathbf{v}_{ss}^{(h)}}{\partial\mathbf{W}}\right\rangle}.
\end{aligned}
\end{equation}
Thus, the gradient is
\begin{equation}
\label{eq: loss function gradient}
    \cfrac{\partial\mathcal{L}}{\partial\mathbf{W}}=\cfrac{2}{k}\sum_{h=1}^{k}{\bigg(\mathbf{v}^{(h)}_{t}-\mathbf{v}^{(h)}_{ss}\bigg)\bigg(\int_{0}^{\infty}{\mathbf{p}\big(\mathbf{v}^{(h)}(t)\big)\mathrm{d}t}\bigg)^{T}}.
\end{equation}

Subsequently, the update of the Laplacian matrix is performed through the following iterative process:
\begin{equation}
\label{eq: PGD}
\mathbf{W}_{s+1}=\text{proj}\bigg[\mathbf{W}_{s}-\eta\bigg(\cfrac{\partial\mathcal{L}}{\partial\mathbf{W}}\bigg|_{\mathbf{W}_{s}}+\beta\mathbf{W}^{T}_{s}\bigg)\bigg],
\end{equation}
for a small enough learning rate, $\eta$.
The process involves computing the gradient of the loss function concerning the Laplacian matrix, followed by a sequence of orthogonal projections. Initially, we project onto the space of matrices with all off-diagonal elements being non-positive, namely $(W_{s+1/3})_{ij}=\min\{0,W_{ij}\}$ for $i\neq j$. 
Then further projects the outcome onto the space of symmetric matrices, namely, $\mathbf{W}_{s+2/3}=(\mathbf{W}_{s+1/3}+\mathbf{W}^{T}_{s+1/3})/2$. 
Finally, we perform a projection to the space of matrices with row sums (or columns) equal to zero, by setting the diagonal element to be minus the sum of all the other elements in the row. The resultant outcome, achieved through these successive projections, represents the optimal direction of progress while adhering to all constraints.

In each iteration of the process, we utilize a numerical solver (e.g., ode45 in Matlab) to solve the dynamic system (\ref{eq: Equation of motion}). Following acquiring the dynamic solution, we calculate the integral presented in expression (\ref{eq: loss function gradient}) through numerical integration. 
This iterative process continues to update the Laplacian matrix until the loss function attains a value lower than a pre-defined tolerance threshold.
This minimization problem outlined in (\ref{eq: optimization}) is addressed in appendix \ref{algorithms} by Algorithm \ref{alg:Global}.
The convergence of iterations is not assured, and criteria for convergence, even in partial form, are beyond the scope of this work. We leave this as an open question for future research. 

\section{Local fine-tuning of the viscous resistances \cite{dillavou2022demonstration,stern2021supervised}}
\label{Local_algo_old}

In the framework of the training protocol, adjustments to the viscous resistances are methodically implemented via an analytical comparison of two discrete flux states, induced within the identical network framework under varying boundary conditions—termed the "free" and "clamped" states. Within the free state, the network aims to fulfill the required task through the implementation of input pressures, denoted by $\mathbf{p_{BC}}$, resulting in the generation of outcome output pressures, $\mathbf{p^{f}_{out}}$, alongside the binary state configurations of each output node. Conversely, the clamped state employs analogous input pressures, $\mathbf{p_{BC}}$, thereby subjecting the output nodes to augmented pressures to instigate a distinct response, $\mathbf{p^{c}_{out}}$.

In cases where the binary state of the output node is acceptable (i.e., meeting the target), the dictated pressure follows the form:
\begin{equation}
\label{eq: learning rule}
    \mathbf{p^{c}_{out}}=\mathbf{p^{f}_{out}}+\eta\big(\mathbf{p^{t}_{out}}-\mathbf{p^{f}_{out}}\big),
\end{equation}
where $\eta\in(0,1)$ represents the amplitude of the nudge towards the desired state, and $\mathbf{p^{t}_{out}}$ is a target pressures vector. 
Subsequently, we posit a learning rule for the tube conductance values
\begin{equation}
\label{eq: delta R}
    \triangle C_{ij}=\frac{\gamma}{2\eta}\big[\big(p_{i}^{f}-p_{j}^{f}\big)^{2}-\big(p_{i}^{c}-p_{j}^{c}\big)^{2}\big],
\end{equation}
where $\gamma$ is a scalar learning rate. 
The learning rule presented in Eq.\,(\ref{eq: delta R}) is distinctly localized, as the change in the conductance of a tube occurs solely in response to the flow through that specific tube. Theoretically, such local learning rules can be practically implemented in physical tubes where the conductance (either radius or length) of the tube is subject to control.

Having achieved the target binary state, clamping the output nodes away from their "free" state towards the desired objective necessitates additional power and induces a lower error state. Small adjustments to the learning degrees of freedom, strategically reducing the energy dissipation of the higher-power clamped state relative to the lower-power free state, lead to a new free-state equilibrium. This new equilibrium manifests with output pressures positioned between the old free and clamped states. 
Analogous to equilibrium propagation, this algorithm approximates global gradient descent in the limit $\eta\ll 1$, allowing the system to learn through iterative updates autonomously \cite{scellier2021deep,dillavou2022demonstration}. Following numerous iterations of the learning rule presented in Eq.\,(\ref{eq: delta R}), the conductance changes tend to approach zero, leading the free state to progressively converge towards the desired target pressures.

\section{Updating the volume vector using an algebraic algorithm}
\label{Algebraic algorithm}
To implement our local learning algorithm, described in section \ref{subsubsec: local supervised learning}, it appears necessary to numerically solve a set of $N$ nonlinear ODE during each iteration of the learning process. For networks that are either extensive or comprised of significant resistances—resulting in slow convergence times—the computational demands can become intricate, inefficient, and potentially non-convergent. We devised an alternative algorithm to optimize the training methodology that facilitates progression through learning iterations without recurrent differential equation resolutions. This approach leverages a knowledge of the bistable network's evolution in response to changes in resistance.
Only in the first iteration, the network's dynamics are calculated by solving the differential equations set, thereby determining the system's equilibrium state—focusing on the volumes, which indicate the binary states of the nodes, although pressures are also considered.
In the next iteration, generally $(s+1)^{th}$, the update of the Laplacian matrix is applied in accordance with rules (\ref{eq: learning rule})-(\ref{eq: delta R}).
Then, we return to Eq.\,(\ref{eq: p_ss_solution}) to iteratively update pressures, $\mathbf{p}_{ss}^{(s+1)}$. 
Following the update of the equilibrium pressure vector, attention is turned to the volume vector update, $\mathbf{v}_{ss}^{(s+1)}$. This phase necessitates differentiating between the system's binary states at the $s^{th}$ step and, based on pressure updates, determining if a node snaped between binary states in the $(s+1)^{th}$ iteration. If a node retains its binary state, the volume vector's adjustment will be small; conversely, a significant modification occurs if the binary state changes (snap-through). The evolution of the constant state volumes vector is subject to a specified rule: 
\begin{equation}
\label{eq: V_ss_s}
    (\mathbf{v_{ss}})_{i}^{(s+1)}=
    \begin{cases}
        f^{-1}_{0}\big((\mathbf{p_{ss}})_{i}^{(s+1)}\big), & \text{if condition 1 hold};\\
        f^{-1}_{1}\big((\mathbf{p_{ss}})_{i}^{(s+1)}\big), & \text{if condition 2 hold};
    \end{cases}
\end{equation}
where $(\mathbf{v_{ss}})_{i}^{(s+1)}$ and $(\mathbf{p_{ss}})_{i}^{(s+1)}$ are the steady state volume and pressure of the $i^{th}$ node in the $(s+1)^{th}$ itteration, respectively. Condition 1 is $\{\{\text{Binary state}=0\cap (\mathbf{p_{ss}})_{i}^{(s+1)}<p_{max}\}\cup\{\text{Binary state}=1\cap (\mathbf{p_{ss}})_{i}^{(s+1)}<p_{min}\}\}$, condition 2 is $\{\{\text{Binary state}=0\cap (\mathbf{p_{ss}})_{i}^{(s+1)}>p_{max}\}\cup\{\text{Binary state}=1\cap (\mathbf{p_{ss}})_{i}^{(s+1)}>p_{min}\}\}$. Moreover, $f_{0}^{-1}(p)$ is the inverse function of $p=f(v)$ where $v<v_{max}$, and $f_{1}^{-1}(p)$ is the inverse function of $p=f(v)$ where $v>v_{min}$. 
Employing these formulas allows the simulation study to advance through mere algebraic calculations—specifically matrix multiplication—thus eliminating the requirement for numerical differential equation solutions in each iteration of the study.

\section{Limitations}
\label{Limitations}
While our research presents significant results, it's important to acknowledge the limitations involved. One such limitation was observed in the context of a 5x5 node lattice network (Fig.\,\ref{fig:Global Metamaterial}). To achieve successful outcomes, we had to permit full connectivity among all nodes, diverging from the conventional layered structure typical of classical neural networks, where each column of nodes connects only to its adjacent columns. This deviation was necessitated because attempts to enforce traditional layered connectivity resulted in the model's failure to converge.

Additionally, the non-convex nature of the optimization model (\ref{eq: optimization}) generated challenges. Given that the pressure-volume characteristics exhibit nonlinearity (via bistability), leading to discontinuities in both the loss function and the error function, the typical convergence graphs were unable to reliably predict the model's convergence or divergence. This complexity underscores the need for further refinement of the model and the development of more robust methods to ensure predictability and stability in the network's learning and adaptation processes.

In the latter part of our study, where networks were trained to achieve specific pressure targets, it was found that networks without a predefined order or structure, i.e., disorder networks (in contrast to those with organized patterns like triangles, squares, or hexagons) performed optimally in training exercises. Furthermore, our observations revealed that networks with nodes connected to 5-6 neighbors performed better than those with fewer or significantly more connections, suggesting an ideal range for connectivity that facilitates effective training.

An additional challenge encountered was the physical limitation imposed on the resistances within the network. Given that resistance values cannot be negative or approach zero without compromising the system's functionality, this constraint inherently limits the network's learning capacity. Consequently, in some instances, our simulations struggled to converge, highlighting a critical area for future investigation. Addressing these limitations requires innovative approaches to network design and learning algorithms, aiming to enhance the adaptability and efficiency of these systems within the bounds of their physical constraints.

\section{Algorithms}
\label{algorithms}
\begin{algorithm}
\caption{Global Supervised Learning of Bistable Flow Network}\label{alg:Global}
\KwIn{Initial conditions- $\mathbf{v}_{0}$,\, Bistable pressure-volume relation- $\mathbf{p(v)}$,\, Initial Laplacian Matrix- $\mathbf{W}_{0}$,\, Set of $k$ volumetric flux- $\{\mathbf{q}^{(h)}\}_{h=1}^{k}$,\, Set of $k$ targets volumes- $\{\mathbf{v}_{t}^{(h)}\}_{h=1}^{k}$,\,Number of nodes- $N$,\, Learning parameter- $\eta$, Regulrization parameter- $\beta$,\, Number of tasks- $k$,\, Learning threshold- $\varepsilon$.}
\KwOut{Laplacian matrix $\mathbf{W}$ satisfing (\ref{eq: optimization}) with threshold error bounded by $\varepsilon$.}
 \textbf{Set:}\, $s=0$,\,$\mathcal{L}=0$,\,$\partial_{\mathbf{W}}\mathcal{L}=\mathbf{0}_{N\times N},\,W^{\text{temp}}=\mathbf{0}_{N\times N}$\;

  \For{$h=1,2,\cdots,k$}
  {
   Solve Eq.\,(\ref{eq: Equation of motion}) by setting $\mathbf{W}=\mathbf{W}_{0}$ with initial conditions $\mathbf{v}_{0}$\;
   Verify that the system has reached a steady state (zero flux through the tubes)\;
   $\mathbf{v}^{(h)}_{ss}\gets$ The volumes obtained in the last time step, i.e., in the steady state\;
    $\mathcal{L}\gets\mathcal{L}+\frac{1}{k}\big|\big|\mathbf{v}^{(h)}_{ss}-\mathbf{v}^{(h)}_{t}\big|\big|_{2}^{2}$\;
    $\partial_{\mathbf{W}}\mathcal{L}\gets \partial_{\mathbf{W}}\mathcal{L}+ \frac{2}{k}\big(\mathbf{v}^{(h)}_{t}-\mathbf{v}^{(h)}_{ss}\big)\big(\int_{0}^{\infty}{\mathbf{p}\big(\mathbf{v}^{(h)}(t)\big)\mathrm{d}t}\big)^{T}$\;
  }

\While{$\mathcal{L}>\varepsilon$}{

   { $\mathbf{W}^{\text{temp}}\gets \mathbf{W}_{s}-\eta\big(\partial_{\mathbf{W}}\mathcal{L}+\beta \mathbf{W}_{s}^{T}\big)$\;
   \For{$i=1,2,\cdots,N$}
{
\For{$j=1,\cdots,N$}
  {
  \If{$i\neq j$}
   {$W_{ij}^{\text{temp}}\gets \text{min}\big\{0,W_{ij}^{\text{temp}}\big\}$\;
   }
   }
    }
  $W^{\text{temp}}\gets \big(W^{\text{temp}}+(W^{\text{temp}})^{T}\big)/2$\;
  \For{$i=1,\cdots,N$}{
  $W_{ii}^{\text{temp}} \gets -\sum_{\substack{j=1 \\ j \neq i}}^{N}W_{ij}^{\text{temp}}$\;
  }
   
  $\mathbf{W}_{s+1}\gets\mathbf{W}^{\text{temp}}$\;
  $[\mathcal{L},\,\partial_{\mathbf{W}}\mathcal{L}]\gets [0,\,\mathbf{0}_{N\times N}]$\;

   \For{$h=1,2,\cdots,k$}
  {
   Solve Eq.\,(\ref{eq: Equation of motion}) by setting $\mathbf{W}=\mathbf{W}_{s+1}$ with initial conditions $\mathbf{v}_{0}$\;
   Verify that the system has reached a steady state (zero flux through the tubes)\;
   $\mathbf{v}^{(h)}_{ss}\gets$ The volumes obtained in the last time step, i.e., in the steady state\;
    $\mathcal{L}\gets\mathcal{L}+\frac{1}{k}\big|\big|\mathbf{v}^{(h)}_{ss}-\mathbf{v}^{(h)}_{t}\big|\big|_{2}^{2}$\;
    $\partial_{\mathbf{W}}\mathcal{L}\gets \partial_{\mathbf{W}}\mathcal{L}+ \frac{2}{k}\big(\mathbf{v}^{(h)}_{t}-\mathbf{v}^{(h)}_{ss}\big)\big(\int_{0}^{\infty}{\mathbf{p}\big(\mathbf{v}^{(h)}(t)\big)\mathrm{d}t}\big)^{T}$\;
  }
  $s\gets s+1$\;
  }
 }
 \Return
 $\mathbf{W}=\mathbf{W}_{s}$\;
\end{algorithm}

\begin{algorithm}
\caption{Local Physical-Supervised Learning of Bistable Flow Network}\label{alg:Local}
\KwData{Initial conditions- $\mathbf{v}_{0}$,\, Bistable pressure-volume relation- $\mathbf{p(v)}$,\, Set of $k$ prescribe inlet pressures- $\{\mathbf{p}_{BC}^{(h)}\}_{h=1}^{k}$,\, Index of inlet nodes- $\mathbf{I_{BC}}$,\, Index of output nodes- $\mathbf{I_{Output}}$,\, Set of $k$ target volumes- $\{\mathbf{v}_{t}^{(h)}\}_{h=1}^{k}$,\, Set of $k$ target pressures- $\{\mathbf{p}_{t}^{(h)}\}_{h=1}^{k}$,\, Set of $k$ target binary states- $\{\text{\textbf{Binary}}^{(h)}\}_{h=1}^{k}$,\, Initial Laplaciam matrix- $\mathbf{W}_{0}$,\, Number of nodes- $N$,\, Number of inlet nodes undergo pressure constrain- $b_{1}$,\, Number of output nodes- $t$,\, Learning parameter- $\eta$,\, Learning rate- $\gamma$,\, Number of tasks- $k$,\, Learning threshold- $\varepsilon$.}
\KwResult{Laplacian matrix $\mathbf{W}$ so that the output nodes’ binary states and corresponding pressures align with the targets with threshold error bounded by $\varepsilon$.}
 \textbf{Set:} $s=0$,\, $\mathcal{L}=0$,\, $\triangle W^{\text{temp}}=\mathbf{0}_{N\times N}$,\,   $\mathbf{v}_{ss}^{f}=\mathbf{0}_{N}$,\,   $\mathbf{p}_{ss}^{f}=\mathbf{0}_{N}$,\, $\mathbf{v}_{ss}^{c}=\mathbf{0}_{N}$,\,   $\mathbf{p}_{ss}^{c}=\mathbf{0}_{N}$,\,  $\mathbf{p}_{out}=\mathbf{0}_{N}$,\,  $\text{\textbf{BS}}^{f}=\mathbf{0}_{N}$,\, $\text{\textbf{BS}}^{c}=\mathbf{0}_{N}$\;

  \For{$h=1,2,\cdots,k$}
  {
   FREE SYSTEM: solve Eq.\,(\ref{eq: Equation of motion}) by setting $\mathbf{W}=\mathbf{W}_{0}$ with initial conditions $\mathbf{v}_{0}$ and boundary conditions $\mathbf{p}_{BC}^{(h)}$ until the system reach a steady state\;
   $[\mathbf{p}^{f,(h)}_{ss},\mathbf{v}^{f,(h)}_{ss}]\gets$ The pressures and volumes obtained in the steady state\;
   $\mathcal{L}\gets\mathcal{L}+\frac{1}{k}\big|\big|\mathbf{v}^{f,(h)}_{ss}-\mathbf{v}^{(h)}_{t}\big|\big|_{2}^{2}$\;

    $[\mathbf{p_{out}^{\text{(\textit{h})}}}$,\,  $\text{\textbf{BS}}^{f,(h)}]\gets$ \,BinaryAndOutPressure($\mathbf{v}^{f,(h)}_{ss},\mathbf{p}^{f,(h)}_{ss},\mathbf{I_{Output}}\, ,\mathbf{Binary},\, \mathbf{p_{t}},\, $\text{\textbf{p(v)}},$N$) according to Algorithm (\ref{alg: function})\;

  CLAMPED SYSTEM: solve Eq.\,(\ref{eq: Equation of motion}) by setting $\mathbf{W}=\mathbf{W}_{0}$ with initial conditions $\mathbf{v}_{0}$ and boundary conditions $\mathbf{p}_{BC}^{(h)}$ in nodes with an index that belongs to $\mathbf{I_{BC}}$ and $\mathbf{p}_{out}^{(h)}$ in nodes with an index that belongs to $\mathbf{I_{output}}$, until the system reach a steady state\;
  $[\mathbf{p}^{c,(h)}_{ss},\mathbf{v}^{c,(h)}_{ss}]\gets$ The pressures and volumes obtained in the steady state\;

    $\text{\textbf{BS}}^{c,(h)}\gets$ \,BinaryAndOutPressure($\mathbf{v}^{c,(h)}_{ss},\mathbf{p}^{c,(h)}_{ss},\mathbf{I_{Output}}\, ,\mathbf{Binary},\, \mathbf{p_{t}},\, $\text{\textbf{p(v)}},$N$) according to Algorithm (\ref{alg: function})\;
        
    $\triangle\mathbf{W}^{\text{temp}}\gets \triangle\mathbf{W}^{\text{temp}}+\frac{\gamma}{2\eta k}\big[\big(\mathbf{p}^{f,(h)}_{ss} \mathbf{1}_{N}^{T}-\mathbf{1}_{N}(\mathbf{p}^{f,(h)}_{ss})^{T}\big)^{2}-\big(\mathbf{p}^{c,(h)}_{ss}\mathbf{1}_{N}^{T}-\mathbf{1}_{N}(\mathbf{p}^{c,(h)}_{ss})^{T}\big)^{2}\big]$, where the square operation is applied element-wise\;

  }

\While{$\mathcal{L}>\varepsilon$}{

   {
   $\mathcal{L}\gets 0$\;
   
   $\mathbf{W}_{s+1}\gets \text{ConstraintLaplacian}(\mathbf{W}_{s}+\triangle\mathbf{W}^{\text{temp}},\, N)$\, according to Algorithm (\ref{alg: function5})\;

\For{$h=1,\cdots,k$}
{

$\mathbf{p}^{f,(h)}_{ss} \gets$\, StadyStatePressureCalc$\big(\mathbf{W_{s+1}},\mathbf{p_{BC}^{\text{\textit{(h)}}}},\mathbf{I_{BC}},b_{1},N\big)$\, according to Algorithm (\ref{alg: function2})\;
$\mathbf{v}^{f,(h)}_{ss} \gets$\, StadyStateVolumeCalc$\big(\mathbf{p}^{f,(h)}_{ss},\mathbf{p(v)},\mathbf{BS}^{f,(h)},N\big)$\, according to Algorithm (\ref{alg: function3})\;

$\mathcal{L}\gets\mathcal{L}+\frac{1}{k}\big|\big|\mathbf{v}^{f,(h)}_{ss}-\mathbf{v}^{(h)}_{t}\big|\big|_{2}^{2}$\;

$[\mathbf{p_{out}^{\text{(\textit{h})}}}$,\,  $\text{\textbf{BS}}^{f,(h)}]\gets$ \,BinaryAndOutPressure($\mathbf{v}^{f,(h)}_{ss},\mathbf{p}^{f,(h)}_{ss},\mathbf{I_{Output}}\, ,\mathbf{Binary},\, \mathbf{p_{t}},\, $\text{\textbf{p(v)}},\, $N$) according to Algorithm (\ref{alg: function})\;

$\mathbf{\hat{p}_{BC}^{\text{\textit{(h)}}}}\gets$\, $\mathbf{p_{BC}^{\text{\textit{(h)}}}}$ in index that belongs to $\mathbf{I_{BC}}$ and $\mathbf{p_{out}^{\text{(\textit{h})}}}$ in index that belongs to $\mathbf{I_{output}}$\;

$\mathbf{p}^{c,(h)}_{ss} \gets$\, StadyStatePressureCalc$\big(\mathbf{W_{s+1}},\mathbf{\hat{p}_{BC}^{\text{\textit{(h)}}}},[\mathbf{I_{BC}};\mathbf{I_{output}}],b_{1}+t,N\big)$\, according to Algorithm (\ref{alg: function2})\;
$\mathbf{v}^{c,(h)}_{ss} \gets$\, StadyStateVolumeCalc$\big(\mathbf{p}^{c,(h)}_{ss},\mathbf{p(v)},\mathbf{BS}^{c,(h)},N\big)$\, according to Algorithm (\ref{alg: function3})\;
$\text{\textbf{BS}}^{c,(h)}\gets$ \,BinaryAndOutPressure($\mathbf{v}^{c,(h)}_{ss},\mathbf{p}^{c,(h)}_{ss},\mathbf{I_{Output}}\, ,\mathbf{Binary},\, \mathbf{p_{t}},\, $\text{\textbf{p(v)}},\, $N$) according to Algorithm (\ref{alg: function})\;

    $\triangle\mathbf{W}^{\text{temp}}\gets \triangle\mathbf{W}^{\text{temp}}+\frac{\gamma}{2\eta k}\big[\big(\mathbf{p}^{f,(h)}_{ss} \mathbf{1}_{N}^{T}-\mathbf{1}_{N}(\mathbf{p}^{f,(h)}_{ss})^{T}\big)^{2}-\big(\mathbf{p}^{c,(h)}_{ss}\mathbf{1}_{N}^{T}-\mathbf{1}_{N}(\mathbf{p}^{c,(h)}_{ss})^{T}\big)^{2}\big]$, where the square operation is applied element-wise\;

}
  
  $s\gets s+1$\;
  }
 }
\end{algorithm}

\begin{algorithm}
\SetAlgoLined
\DontPrintSemicolon
\KwIn{Laplacian matrix- $\mathbf{W}$,\, Number of nodes- $N$.}    
\KwOut{Constraint Laplacian matrix-
$\mathbf{\hat{W}}$.}

    \SetKwFunction{FMain}{ConstraintLaplacian}
    \SetKwProg{Fn}{Function}{:}{}
    \Fn{\FMain{$\mathbf{W},N$}}{
    \textbf{Set:}   $\mathbf{\hat{W}=0_{N\times N}}$;\;

   \For{$i=1,2,\cdots,N$}
{
\For{$j=i+1,\cdots,N$}
  {
   $\hat{W}_{ij}\gets \text{min}\big\{0,W_{ij}\big\}$\;
   $\hat{W}_{ji}\gets \hat{W}_{ji}$
   }
    }

  \For{$i=1,\cdots,N$}{
  $\hat{W}_{ii} \gets -\sum_{\substack{j=1 \\ j \neq i}}^{N}\hat{W}_{ij}$\;
  }
  }
\Return $\mathbf{\hat{W}}$;
      
\textbf{End Function}
\caption{The Constraint Laplacian Matrix}
\label{alg: function5}
\end{algorithm}

\begin{algorithm}
\SetAlgoLined
\DontPrintSemicolon
\KwIn{Volumes- $ \mathbf{v}$,\, Pressures- $ \mathbf{p}$,\, Index of output nodes- $\mathbf{I_{Output}}$,\, Target binary states- $\text{\textbf{Binary}}$,\, Bistable pressure-volume relation- $\text{\textbf{p(v)}}$,\, Target pressures- $\mathbf{p_{t}}$,\, Number of nodes- $N$.}    
\KwOut{Constraint output pressures- 
$\mathbf{p_{out}}$,\, Binary states of the nodes- $\mathbf{BS}$.}

    \SetKwFunction{FMain}{BinaryAndOutPressure}
    \SetKwProg{Fn}{Function}{:}{}
    \Fn{\FMain{$\mathbf{v},\mathbf{p},\mathbf{I_{Output}},\mathbf{Binary},\mathbf{p_{t}},\, \text{\textbf{p(v)}},N$}}{
    \textbf{Set:}   $\mathbf{p}_{out}=\mathbf{0}_{N}$,\,  $\text{\textbf{BS}}=\mathbf{0}_{N}$,\, $\text{count}=1$;\;
    Calculate $(p_{\max},\, p_{\min},\, v_{\max},\, v_{\min})$ from bistable pressure-volume relation- $\mathbf{p(v)}$;\;
    
       \For{$i=1,\cdots,N$}
{     \If{$v_{i}>v_{\min}$}{
   $\text{BS}_{i}\gets1$;\;
   }}
   
   \For{$i\in\mathbf{I_{Output}}$}{

    \eIf{$\text{BS}_{i}=\text{Binary}_{\text{count}}$}
    {
    $p_{out,i}\gets p_{i}+\eta({p}_{t,\text{count}}- p_{i})$;\;
    }{
    \eIf{$\text{BS}_{i}=0$}
    {
    $p_{out,i}\gets \alpha_{1} p_{\max}\,$ where $\alpha_{1}>1$;\;}
    {$p_{out,i}\gets \alpha_{2} p_{\min}\,$ where $0<\alpha_{2}<1$;\;}
    }
    $\text{count}\gets\text{count}+1$;\;
   }

        \textbf{return:} $\mathbf{p_{out}}$,\,  $\text{\textbf{BS}}$;\;
}
\textbf{End Function}
\caption{Finding A Binary State And Determining Constraint Output Pressures}
\label{alg: function}
\end{algorithm}

\begin{algorithm}
\SetAlgoLined
\DontPrintSemicolon
\KwIn{Laplacian matrix- $\mathbf{W}$,\, Pressures in boundary condition- $\mathbf{p_{BC}}$,\, Index of boundary condition nodes- $\mathbf{I_{BC}}$,\, Number of inlet nodes- $b$,\, Number of nodes- $N$.}    
\KwOut{Nodes' pressures at steady states-
$\mathbf{p_{ss}}$.}

    \SetKwFunction{FMain}{StadyStatePressureCalc}
    \SetKwProg{Fn}{Function}{:}{}
    \Fn{\FMain{$\mathbf{W},\mathbf{p_{BC}},\mathbf{I_{BC}},b,N$}}{
    \textbf{Set:}   $\mathbf{\hat{W}=0_{N\times N}}$;\;

    \For{$i = 1:b$}
{$\mathbf{\hat{W}} \gets$\, Swap rows '$i$' and '$(\mathbf{I_{BC}})_{i}$' in $\mathbf{W};$\;
$\mathbf{\hat{W}} \gets$\, Swap columns '$i$' and '$(\mathbf{I_{BC}})_{i}$' in $\mathbf{\hat{W}};$\;
}
  
$\mathbf{\hat{p}}_{ss}\gets [\mathbf{p_{BC}};\,  -\mathbf{\hat{W}_{33}}^{-1}\mathbf{\hat{W}_{31}}\mathbf{p_{BC}}]$ according to Eq.\,(\ref{eq: p_ss_solution});\;

\For{$i = 1:b$}
{$\mathbf{p}_{ss} \gets$\, Swap rows '$i$' and '$(\mathbf{I_{BC}})_{i}$' in $\mathbf{\hat{p}}_{ss};$\;
}
}

\Return $\mathbf{p}_{ss}$;
      
\textbf{End Function}
\caption{Stady-State Pressure Calculation - According to Eq.\,(\ref{eq: p_ss_solution})}
\label{alg: function2}
\end{algorithm}

\begin{algorithm}
\SetAlgoLined
\DontPrintSemicolon
\KwIn{steady state pressures- $\mathbf{p_{ss}}$,\, Bistable pressure-volume relation- $\text{\textbf{p(v)}}$,\, Binary states of nodes- $\textbf{BS}$,\,  Number of nodes- $N$.}    
\KwOut{Nodes' volumes at steady states-
$\mathbf{v_{ss}}$.}

    \SetKwFunction{FMain}{StadyStateVolumeCalc}
    \SetKwProg{Fn}{Function}{:}{}
    \Fn{\FMain{$\mathbf{p_{ss}},\mathbf{p(v)},\mathbf{BS},N$}}{
    \textbf{Set:}   $\mathbf{v_{ss}=0_{N}}$;\;
    Calculate $(p_{\max},\, p_{\min},\, v_{\max},\, v_{\min})$ from bistable pressure-volume relation- $\mathbf{p(v)}$;\;
    Calculate the invers function $f_{0}^{-1}(p)$ where $v<v_{\max}$;\;
    Calculate the invers function $f_{1}^{-1}(p)$ where $v>v_{\min}$;\; 
\For{$i=1,\cdots,N$}
{
\eIf{$\{BS_{i}=0\And p_{ss,i}<p_{max}\}\, \text{or}\,  \{BS_{i}=1\And p_{ss,i}<p_{min}\}$}
{$v_{ss,i}\gets f_{0}^{-1}\big(p_{ss,i}\big)$;\;}
{$v_{ss,i}\gets f_{1}^{-1}\big(p_{ss,i}\big)$;\;}

}
}

\Return $\mathbf{v}_{ss}$;
      
\textbf{End Function}
\caption{Stady-State Volumes Calculation - According to Eq.\,(\ref{eq: V_ss_s})}
\label{alg: function3}
\end{algorithm}

\newpage
\bibliographystyle{unsrtnat}
\bibliography{references}  



\end{document}